\documentclass{article} % For LaTeX2e
\usepackage{iclr2027_conference,times}

\usepackage{amsmath,amsfonts,bm}

\def\eqref#1{equation~\ref{#1}}
\def\1{\bm{1}}

\DeclareMathAlphabet{\mathsfit}{\encodingdefault}{\sfdefault}{m}{sl}
\SetMathAlphabet{\mathsfit}{bold}{\encodingdefault}{\sfdefault}{bx}{n}

\usepackage{hyperref}
\usepackage{url}
\usepackage{float}
\usepackage{amssymb}
\usepackage{booktabs}
\usepackage{pifont}
\usepackage{graphicx}
\usepackage{makecell}
\usepackage{wrapfig}
\usepackage{xcolor}
\usepackage{multirow}

\newcommand{\cmark}{\textcolor{green!60!black}{\ding{51}}}
\newcommand{\xmark}{\textcolor{red!80!black}{\ding{55}}}
\newcommand{\hdr}[2]{\textbf{\begin{tabular}[b]{@{}c@{}}#1\\#2\end{tabular}}}
\newcommand{\hdro}[1]{\raisebox{\dimexpr(\ht\strutbox+\dp\strutbox)/2\relax}{\textbf{#1}}}
\newcommand{\pmark}{$\triangle$}

\title{AnchorReasoning: A Visual Grounding and Causal Reasoning Dataset in Long-Tail Autonomous Driving Scenarios}

\author{Zhipeng Bao$^{1}$, Wenjie Zhao$^{1}$, Tianle Zhu$^{1}$, Haohua Que$^{1}$, \\
\textbf{Chence Yang$^{2}$, Geng Yuan$^{2}$, Qianwen Li$^{1}$\thanks{Corresponding author.}} \\
{\normalfont $^{1}$School of Environmental, Civil, Agricultural and Mechanical Engineering, University of Georgia} \\
{\normalfont $^{2}$School of Computing, University of Georgia}
}

\iclrfinalcopy % Non-anonymous version for arXiv (removes line numbers and anonymization).
\begin{document}

\maketitle
% arXiv version: no ICLR running header
\lhead{}
\renewcommand{\headrulewidth}{0pt}

\begin{abstract}
Vision-language models (VLMs) offer a promising approach to long-tail autonomous driving, but existing driving datasets provide limited supervision for connecting decision-critical visual evidence with reasoning and planning. We introduce AnchorReasoning, a visually grounded reasoning dataset built on WOD-E2E, containing 416,119 annotated frames and 395,379 decision-critical elements across four major categories and 19 fine-grained types. Each frame is organized as a visually grounded chain-of-thought (VG-CoT) that links decision-critical element identification and localization, element attributes and implications, driving-action rationale, and action and trajectory planning. We further develop a curriculum supervised fine-tuning strategy that progressively learns these hierarchical capabilities, together with an object-size-aware grounding metric for evaluating localization quality. Experiments across eight general-purpose, embodied-AI, and AV-specific backbones show that VG-CoT supervision improves grounded reasoning and trajectory prediction. Across models, 5-s ADE and FDE decrease by 7.84 and 11.86, while RFS Frame and Cluster improve by 1.66 and 1.70. These gains are achieved with 18.5 fewer reasoning tokens and 0.32 s/frame lower inference latency on average, demonstrating the value of visually grounded, decision-focused supervision for VLM reasoning and planning in long-tail autonomous driving.
\end{abstract}

\section{Introduction}

Recent autonomous driving research has increasingly focused on addressing long-tail scenarios. These rare and complex situations often require a deeper understanding of the driving context and deliberate reasoning to derive appropriate behavioral plans. Vision-language models (VLMs), with their strong multimodal understanding and commonsense reasoning capabilities, have therefore emerged as a promising paradigm for handling such challenging scenarios \citep{wang2025alpamayo,zhou2026autovla}. However, effectively harnessing these capabilities for long-tail driving largely depends on how the training data are organized.

Early VLM-oriented driving datasets were primarily organized around scene descriptions or captioning annotations \citep{kim2018textual,liu2023traffic}. Such annotations provide holistic supervision of the driving environment, enabling models to learn high-level scene semantics and contextual information. However, coarse scene-level descriptions provide limited supervision for fine-grained, object-centric understanding. More importantly, they provide little explicit guidance on how scene observations should be translated into driving decisions.

\begin{table}[b]
\centering
\caption{\textbf{Comparison of representative VLM-oriented driving datasets.}}
\label{tab:dataset_comparison}

\scriptsize
\setlength{\tabcolsep}{1.2pt}
\renewcommand{\arraystretch}{0.95}

\begin{tabular*}{\textwidth}
{@{\extracolsep{\fill}}lcccccccc@{}}
\toprule

& & & &
\multicolumn{3}{c}{\textbf{Decision-critical Elements}}
& & \\
\cmidrule(lr){5-7}

\hdro{Dataset}
& \hdr{\# Ann.}{Frames}
& \hdro{Long-tail}
& \hdr{Decision-}{critical Element}
& \hdr{Visual}{Grounding}
& \hdr{Impact}{Rank}
& \hdro{Implication}
& \hdro{Reasoning}
& \hdr{Action /}{Trajectory}
\\
\midrule

nuScenes-QA \citep{qian2024nuscenes}
& 34,149
& \xmark
& \xmark
& \xmark
& \xmark
& \xmark
& QA
& -- \\

LingoQA \citep{marcu2024lingoqa}
& 28K clips
& \xmark
& \cmark
& \xmark
& \xmark
& \xmark
& QA
& Action \\

Rank2Tell \citep{sachdeva2024rank2tell}
& 3,833
& \xmark
& \cmark
& \cmark
& \cmark
& \xmark
& QA
& -- \\

DriveLM \citep{sima2024drivelm}
& 4,871
& \xmark
& \cmark
& \xmark
& \cmark
& \cmark
& Graph QA
& Both \\

DriveLMM-o1 \citep{ishaq2025drivelmm}
& 2,501
& \xmark
& \cmark
& \xmark
& \xmark
& \xmark
& CoT
& Action \\

STRIDE-QA \citep{ishihara2026stride}
& 270K
& \xmark
& \xmark
& \xmark
& \xmark
& \xmark
& Grounded QA
& -- \\

CoVLA \citep{arai2025covla}
& 6.0M
& \xmark
& \xmark
& \xmark
& \xmark
& \xmark
& Description
& Both \\

nuReasoning \citep{huang2026nureasoning}
& $\sim$3.78M est.
& \cmark
& \cmark
& \pmark
& \xmark
& \cmark
& QA \& CoT
& Both \\

\midrule

\textbf{AnchorReasoning (Ours)}
& \textbf{416K}
& \cmark
& \cmark
& \cmark
& \cmark
& \cmark
& \textbf{VG-CoT}
& \textbf{Both} \\

\bottomrule
\end{tabular*}

\vspace{2pt}

\begin{minipage}{\textwidth}
\scriptsize
\textit{Note.}
``Graph QA'' organizes QA pairs as a dependency graph linking perception,
prediction, and planning questions.
``Grounded QA'' associates QA pairs with corresponding visual regions or objects.
$\triangle$ denotes partial grounding: nuReasoning grounds objects in spatial
reasoning with 2D image locations, but does not explicitly propagate this
grounding into decision reasoning or trajectory generation.
\end{minipage}
\vspace{-0.8em}
\end{table}

To provide more targeted supervision, recent datasets have increasingly adopted question-answering (QA) and chain-of-thought (CoT) annotations \citep{wu2025language,marcu2024lingoqa,qian2024nuscenes}. QA-based datasets decompose driving scenes into object-level questions, allowing models to learn individual visual concepts. Such supervision enables more fine-grained probing and learning of driving implications of individual elements. However, QA-based supervision primarily trains VLMs to answer isolated questions, providing limited guidance on how multiple pieces of evidence are connected into a coherent and structured reasoning process \citep{nie2024reason2drive}. CoT-based annotations mitigate this limitation by modeling the intermediate reasoning steps, enabling VLMs to learn a coherent driving reasoning chain that connects scene observation, contextual interpretation, and decision-making \citep{ishaq2025drivelmm,arai2025covla}. Despite these advances, most existing CoT supervision does not prioritize decision-critical evidence. Consequently, VLMs may incorporate decision-irrelevant elements into their reasoning, resulting in verbose or distracted outputs that compromise both quality and efficiency~\citep{qian2025agentthink}. In addition, current CoT annotations are predominantly textual and rarely ground critical elements to their corresponding visual regions. This missing connection may cause VLMs to rely on language priors rather than the actual scene, producing interpretations that are weakly supported by visual observations and increasing the risk of hallucination \citep{xie2025vlms,chang2026probing}. Moreover, most existing datasets are dominated by routine driving scenarios and provide limited coverage of long-tail situations. 

To address these limitations, we introduce AnchorReasoning, a visual grounding and causal reasoning dataset for VLMs in long-tail driving scenarios (Table~\ref{tab:dataset_comparison}). Built upon WOD-E2E \citep{xu2026wod}, AnchorReasoning augments real-world long-tail driving clips with decision-critical visual grounding annotations and driving reasoning chains tailored for VLM training. For each frame, the annotations are organized as a visually grounded chain-of-thought (VG-CoT): observation and grounding, element implication, driving action rationale, and driving action and trajectory planning. First, it guides VLMs to identify and localize decision-critical elements, such as roadway users and traffic control devices. It then enables VLMs to interpret element attributes and infer their implications for ego driving. VLMs then consolidate multiple implications to derive the action plan and future trajectory, while providing a rationale. In addition, we develop a curriculum supervised fine-tuning (SFT) strategy to fully leverage the proposed dataset. The curriculum progressively equips VLMs with visually grounded causal reasoning and planning capabilities through the VG-CoT supervision. We further design a point-based visual-grounding metric to evaluate the localization accuracy of predicted decision-critical elements. Experimental results across different backbones demonstrate that VG-CoT supervision equips models with visual grounding capability, enabling them to incorporate visual evidence into higher-quality reasoning and more accurate trajectory prediction. It improves the Rater Feedback Score (RFS) Frame and Cluster by 1.66 and 1.70. These gains are achieved with 18.5 fewer reasoning tokens on average and a 0.32\,s/frame reduction in inference latency. The main contributions of this paper are summarized as follows: 

\begin{figure}[t]
    \centering
    \includegraphics[width=\linewidth]{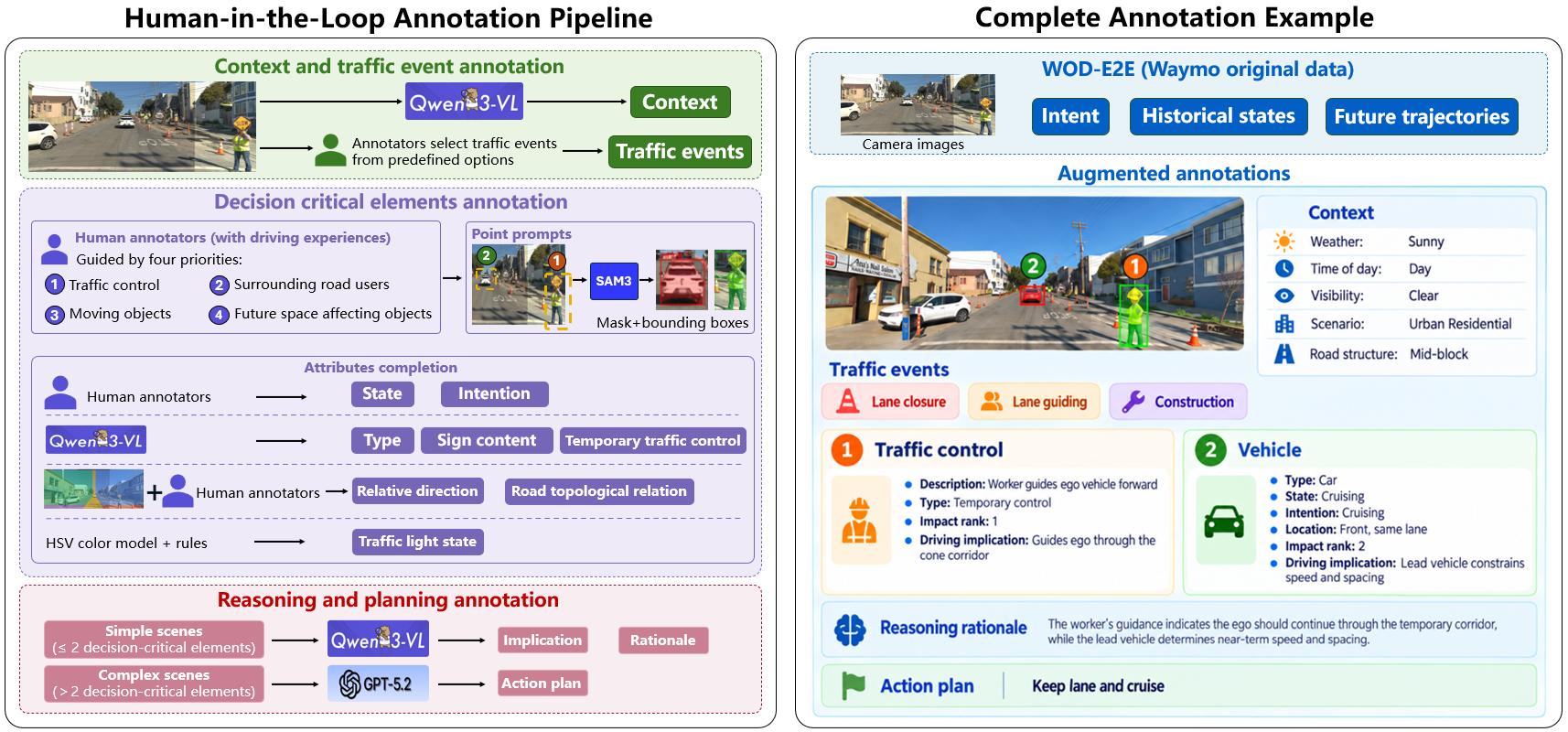}
    \caption{AnchorReasoning annotation pipeline, and example.}
    \label{main-fig}
\end{figure}

\begin{itemize} 
\item We introduce AnchorReasoning, a long-tail driving dataset for VLM training that provides decision-relevant visual grounding, fine-grained element attributes and semantics, and structured reasoning annotations. These annotations enable models to identify, localize, and correctly interpret the elements that truly influence driving decisions, thereby improving their trajectory planning performance in long-tail autonomous driving scenarios.

\item We propose a curriculum SFT strategy that follows the structure of VG-CoT, allowing models to fully leverage the hierarchical supervision and progressively acquire visual grounding, element implication, driving action rationale, behavior planning, and trajectory generation capabilities.

\item We introduce an object-size-aware visual grounding metric that evaluates predicted grounding points using scale-adaptive point-to-mask distances, providing a consistent measure of localization quality across elements of different sizes.
\end{itemize}

\section{Related Work}
\paragraph{Language Description and Explanation.}
Early language-supervised driving datasets focus on describing driving behavior and explaining action decisions. BDD-X provides natural-language descriptions of driving actions and their justifications \citep{kim2018textual}, while DRAMA further introduces localization of risk-relevant objects together with hazard explanations \citep{malla2023drama}. These works establish language-based supervision for driving scene understanding and behavior explanation.

\paragraph{Question Answering.}
VQA-based datasets provide more structured supervision through targeted questions about driving scenes. NuScenes-QA covers object existence, attributes, states, and spatial relationships \citep{qian2024nuscenes}; LingoQA extends the scope to localization, anticipation, attention, and action justification \citep{marcu2024lingoqa}; and Rank2Tell identifies and ranks decision-relevant traffic participants with natural-language explanations \citep{sachdeva2024rank2tell}. DriveLM further introduces Graph VQA, connecting perception, prediction, and planning questions through logical dependencies \citep{sima2024drivelm}. These datasets support fine-grained learning of object- and task-level driving knowledge.

\paragraph{Chain-of-Thought Reasoning.}
Recent datasets move toward structured reasoning chains for driving. Reason2Drive introduces reasoning across perception, prediction, and decision-making \citep{nie2024reason2drive}, while DriveLMM-o1 provides step-by-step rationales for perception, prediction, and planning \citep{ishaq2025drivelmm}. nuReasoning further extends reasoning supervision to real-world long-tail scenarios with spatial, decision, and counterfactual reasoning \citep{huang2026nureasoning}.

\begin{figure}[htbp]
    \centering
    \includegraphics[width=0.8\linewidth]{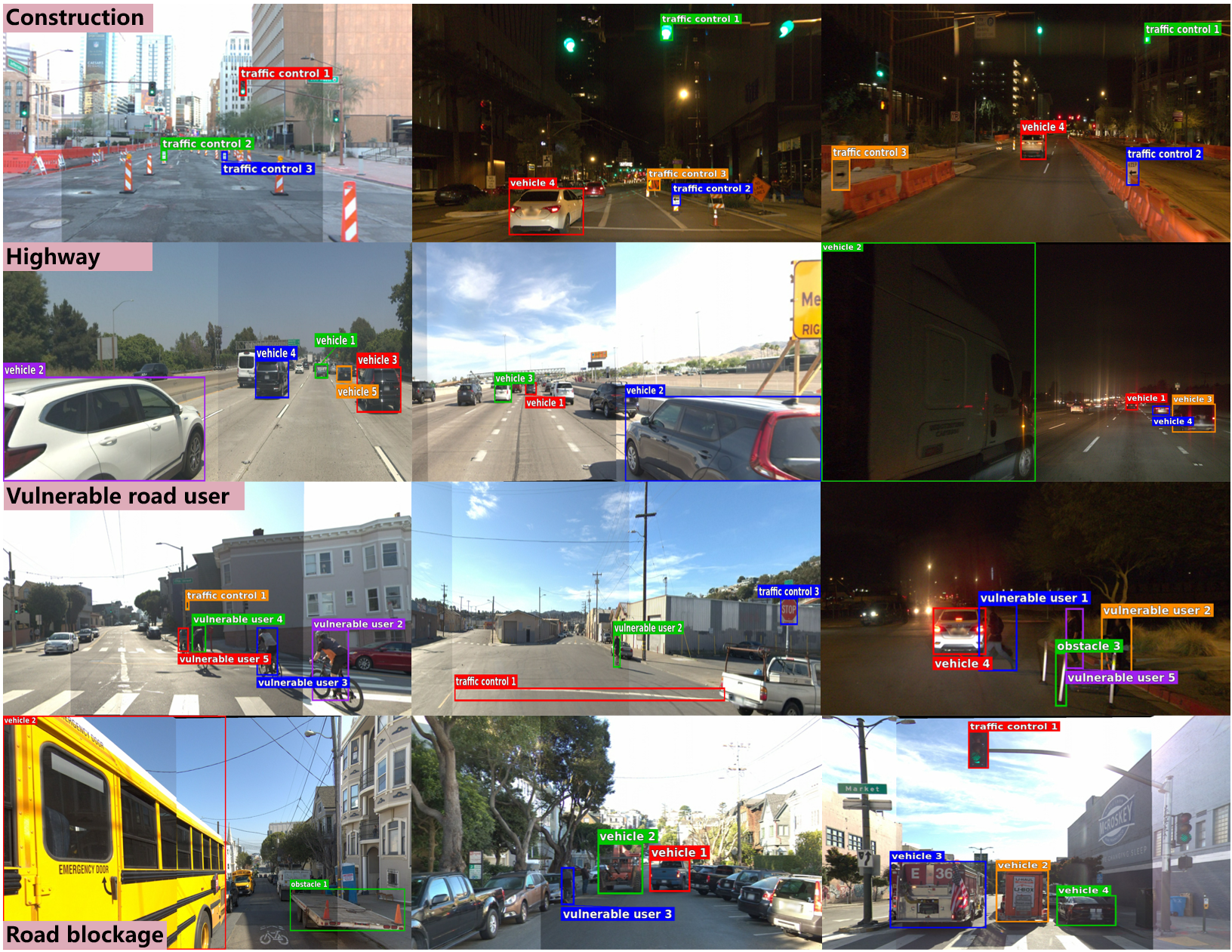}
    \caption{Representative annotation examples across diverse driving scenarios. Decision-critical elements are highlighted with bounding boxes and category-specific labels.}
    \label{fig:visual-samples}
\end{figure}

\section{AnchorReasoning Dataset}
\subsection{Overview} 
AnchorReasoning is a visually grounded reasoning dataset designed to train and evaluate VLM in long-tail scenarios. Built upon WOD-E2E, it augments real-world long-tail driving clips with annotations centered on decision-critical elements. AnchorReasoning identifies and visually grounds the elements that influence the ego driving, and further annotates their attributes and implications. These annotations are organized into a VG-CoT that connects grounding, scene understanding, causal reasoning, action and trajectory prediction. The dataset contains over \textbf{11} hours of driving data across \textbf{2492} clips, comprising \textbf{416,119} annotated frames and \textbf{395,379} decision-critical elements.

\subsection{VG-CoT Annotation Schema}

\begin{wrapfigure}{r}{0.38\textwidth}
    \vspace{-1.0em}
    \centering
    \includegraphics[width=\linewidth]{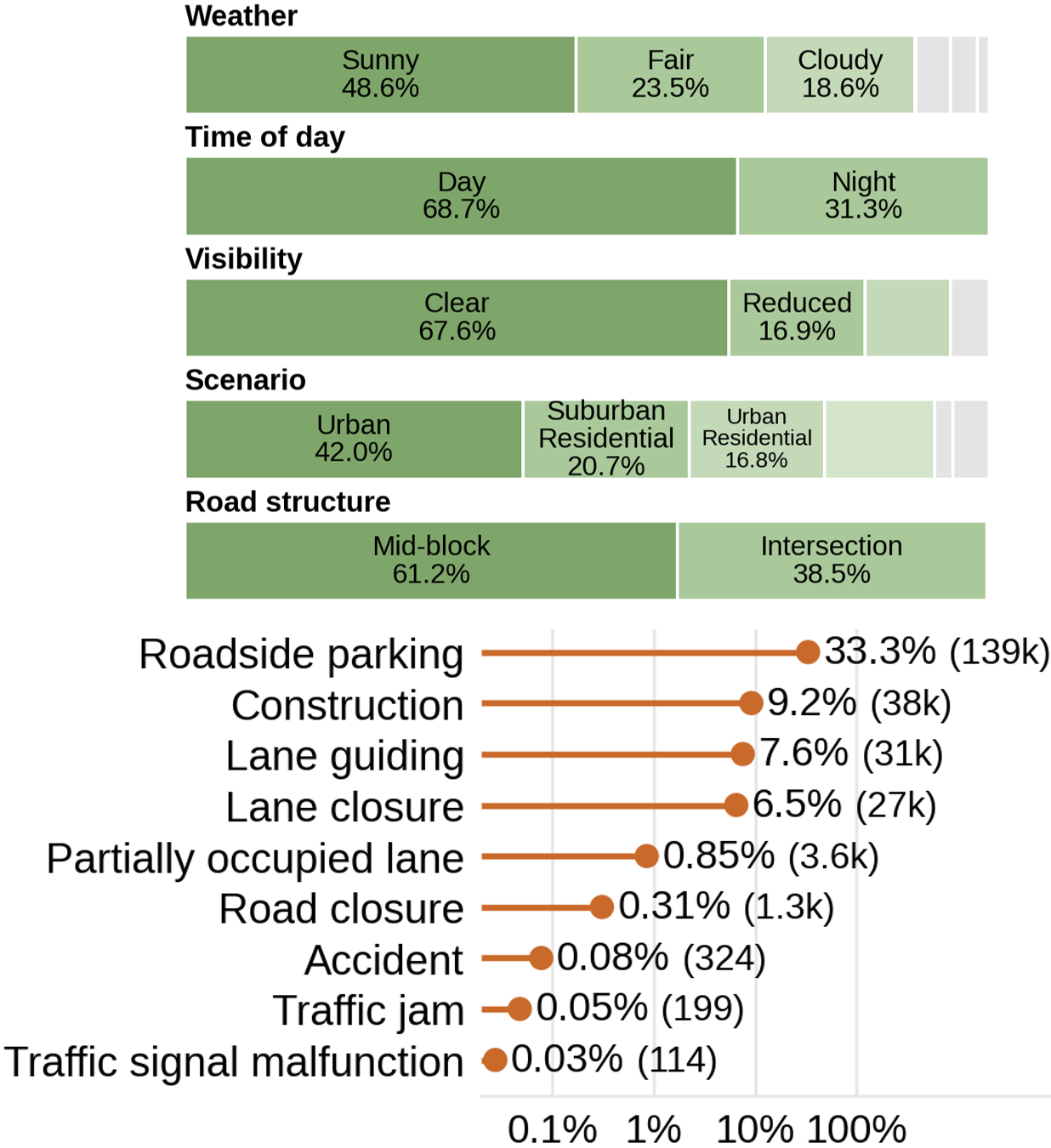}
    \caption{Context and traffic event distribution. Not labeled: Weather: Rainy 4.4\%, Unknown 3.4\%, Fog 1.5\%; Visibility: Limited 10.6\%, Poor 4.9\%; Scenario: Suburban 13.7\%, Urban Highway 2.3\%, Other 4.5\%; Road structure: Other 0.3\%.}
    \label{fig:scene}
    \vspace{-0.5em}
\end{wrapfigure}

VG-CoT is designed to guide VLMs through a hierarchical and evidence-grounded driving reasoning process. It encourages models to reason from the observed scene instead of relying primarily on language priors, thereby improving reasoning fidelity and reducing hallucinations. VG-CoT first provides scene-level context and event-level traffic information. The context annotations describe global driving conditions (Fig.~\ref{fig:scene}), providing the background needed to interpret appropriate driving behavior under different environments. Traffic event annotations further capture events such as roadside parking and construction. These annotations provide event-level constraints that may not be fully represented by individual objects.
   
The core of VG-CoT is the annotation of decision-critical elements, which directs the model toward visual evidence that influences the current driving decision (Fig. ~\ref{fig:visual-samples}). Each element is annotated with both visual grounding (bounding boxes and segmentation masks), and structured semantic information. We group decision-critical elements into four major categories: vehicles, vulnerable road users (VRUs), obstacles, and traffic-control elements. For vehicles, VRUs, and obstacles, the annotated attributes include type, location, state, intention, and impact rank. Location describes both the relative direction of the element and its road-topological position. State represents the currently observed condition, whereas intention describes the likely short-term behavior, such as turning, or crossing the road. Impact rank indicates the relative priority of each element in the current decision, where a smaller rank number denotes a stronger influence on ego driving. This ranking guides the model to allocate attention among multiple critical elements in a human-like manner. For traffic-control elements, other than the impact rank, the annotations capture information such as traffic-light states, sign semantics, and descriptions of temporary traffic control. Each decision-critical element is further associated with an implication, which explains how the observed element may affect the ego vehicle and what action responses are appropriate. This encourages the model to reason about one critical factor at a time before integrating multiple factors, thereby reducing interference from unrelated information and facilitating the subsequent global reasoning process.

Finally, the reasoning and planning stage consolidates multiple element-level implications to derive the final driving decision. The rationale weighs these cues according to their relevance and impact, producing longitudinal and lateral actions. Because high-level actions can serve as an intermediate representation between semantic reasoning and continuous trajectory, AnchorReasoning includes both action annotations and future trajectories, supporting models that predict either meta-actions or trajectories directly.

\begin{figure}[htbp]
    \centering
    \includegraphics[width=\linewidth]{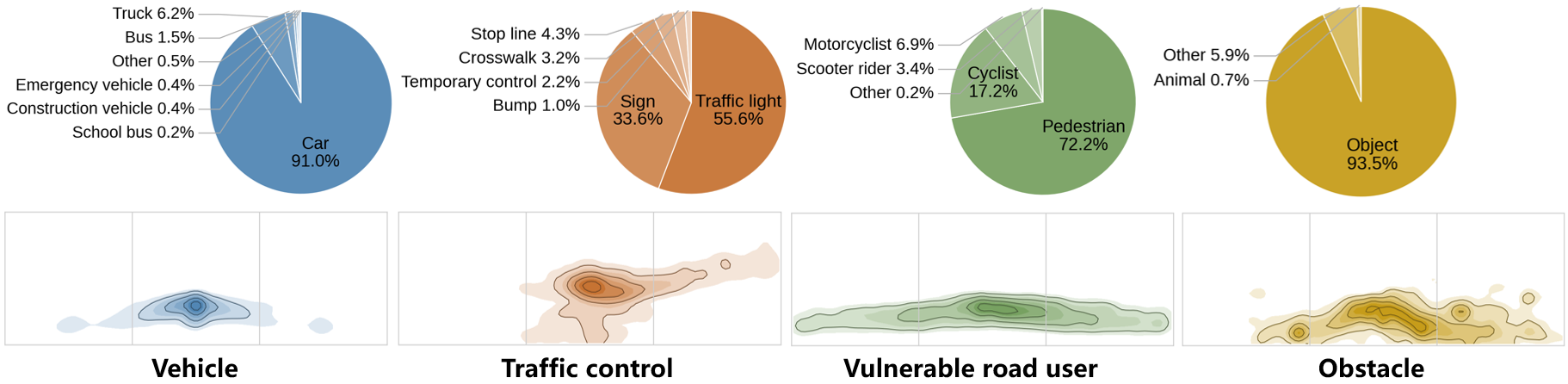}
    \caption{Decision-critical element composition and spatial distribution.}
    \label{fig:elements}
\end{figure}

\subsection{Human-in-the-Loop Annotation and Quality Control}

\begin{wrapfigure}{r}{0.3\textwidth}
    \vspace{-0.8em}
    \centering
    \includegraphics[width=\linewidth]{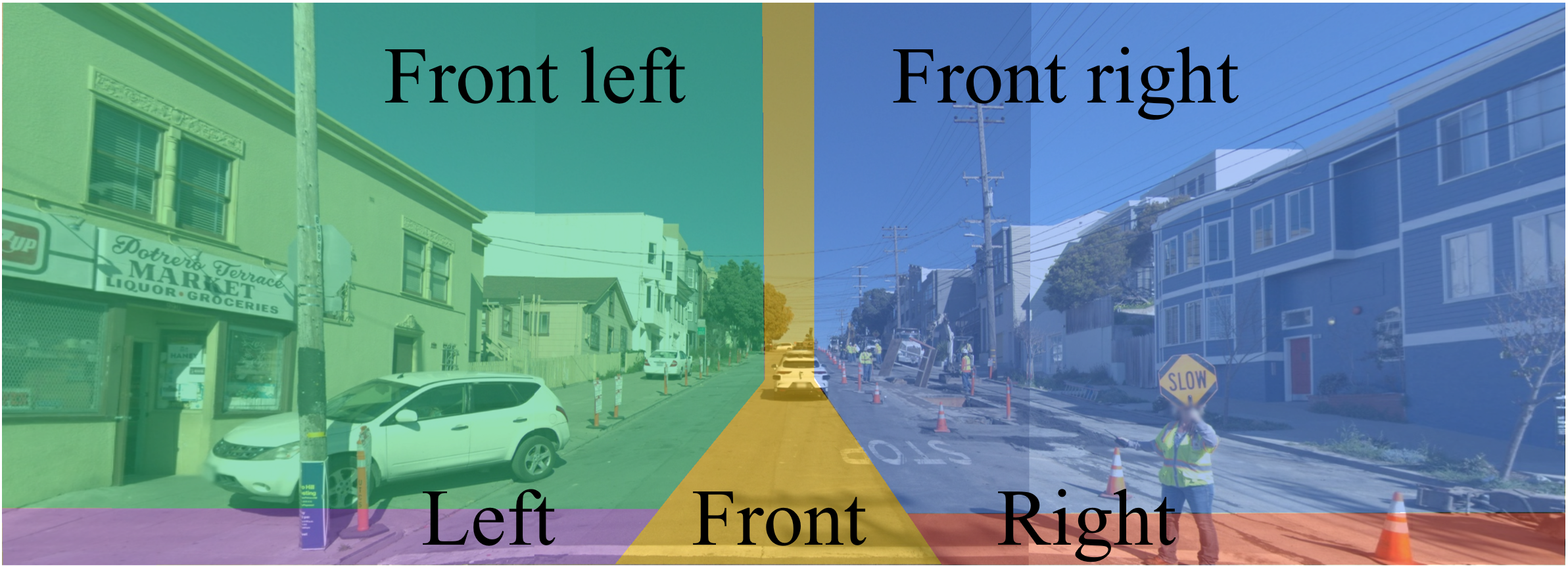}
    \caption{Predefined spatial regions.}
    \label{fig:region}
    \vspace{-0.6em}
\end{wrapfigure}

We adopt a hybrid human-in-the-loop annotation pipeline that combines human judgment, VLM-assisted annotation, promptable segmentation, and rule-based methods. At the scene level, Qwen3-VL generates contextual labels, while human annotators identify traffic events. For decision-critical elements, annotators with driving experience first identify and rank the elements they would prioritize under the same driving situation. To improve consistency, they follow four guidelines: prioritizing traffic-control elements, nearby elements, moving elements, and elements likely to affect the ego vehicle’s future driving space. At most five decision-critical elements are retained per frame to keep the supervision concise. Annotators then provide point prompts for each selected element, which are passed to SAM3 to generate instance masks and corresponding bounding boxes.

\begin{wrapfigure}{r}{0.42\textwidth}
    \vspace{-0.8em}
    \centering
    \includegraphics[width=\linewidth]{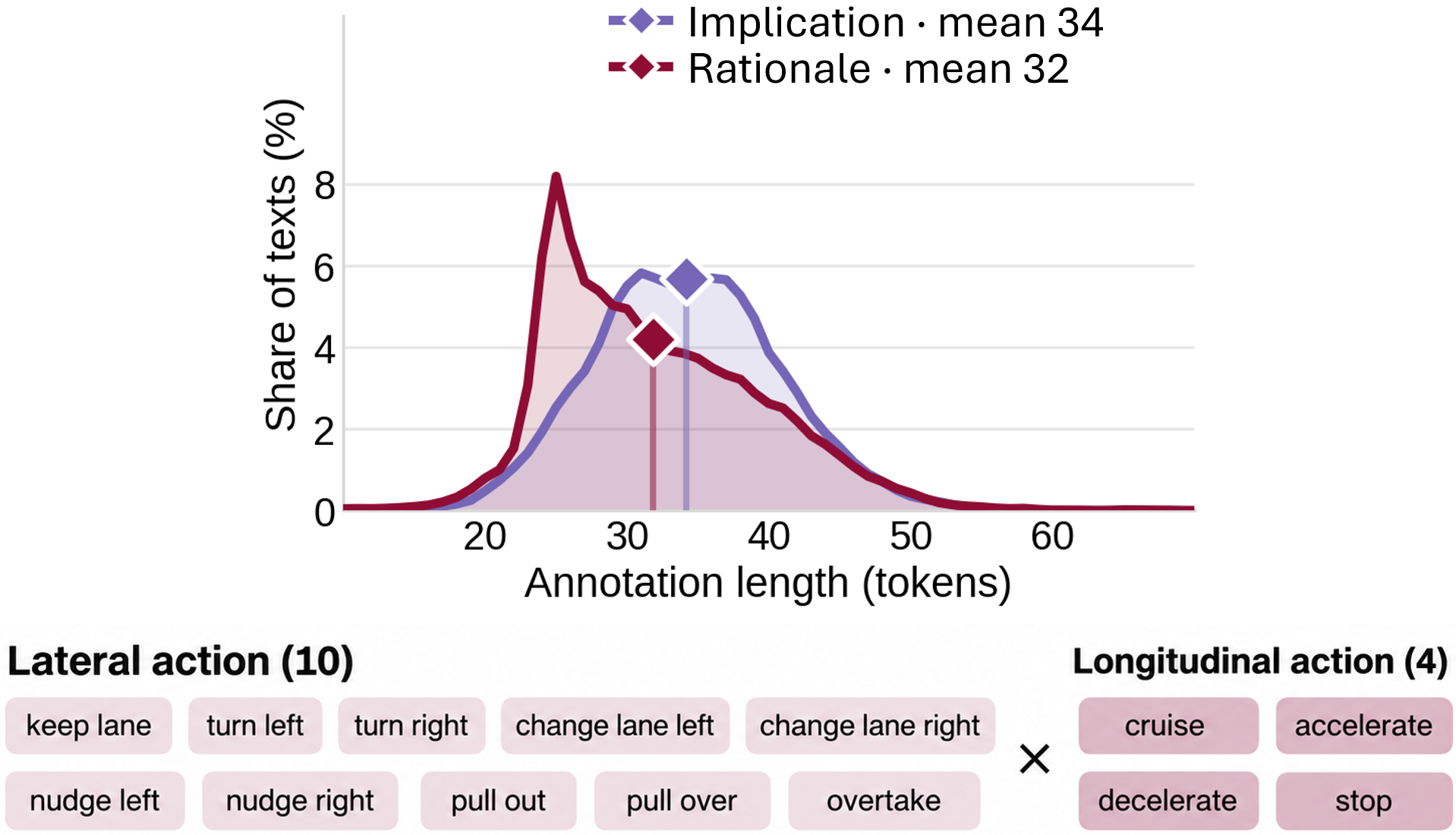}
    \caption{Free-text annotation length and action space.}
    \label{fig:annotation_action}
    \vspace{-0.8em}
\end{wrapfigure}

Element attributes are subsequently obtained through human and automatic annotation. Human annotators provide element states and short-term intentions, while Qwen3-VL generates element types, sign content, and temporary traffic-control descriptions. Element location combines manually annotated road-topological relations with zone-based relative directions inferred from bounding-box positions and their overlap with predefined regions (Fig. ~\ref{fig:region}). Traffic-light states are determined using Hue–Saturation–Value-based rules within the detected traffic-light regions (Appendix ~\ref{app:traffic_light}).

Based on these grounded element annotations, implications are generated using a complexity-aware strategy. Scenes with no more than two decision-critical elements are processed using a locally deployed Qwen3-VL, whereas scenes with richer interactions are handled by GPT-5.2. The same strategy is applied to the final reasoning and planning stage. This tiered pipeline can balance annotation quality and cost.

\paragraph{Quality Control.}

Before large-scale annotation, each automatic annotation tool is evaluated to verify its reliability; the evaluation method and quantitative results are provided in Appendix~\ref{app:auto_annotation}. After annotation is completed, we employ a multi-stage quality-control procedure. For context and traffic event annotations, 50\% of the samples are randomly reviewed and corrected by human annotators. Decision-critical element selection and impact ranking are further verified through cross-annotation review. Each sample is checked by a second annotator, and substantial disagreements are resolved by a third annotator through majority voting. For long-form annotations, including implications, rationales, and action plans, recurrent error patterns are summarized and used to refine the system prompts, after which the corresponding annotations are regenerated. We additionally apply a rule-based consistency check between action plans and future trajectories to enforce expected geometric and kinematic consistency (Appendix ~\ref{app:conssistency}). Inconsistent samples are flagged for manual review and correction.

\subsection{Dataset Statistics}
The dataset contains approximately \textbf{395K} decision-critical elements across \textbf{4} high-level categories and \textbf{19} fine-grained element types, providing extensive object-level supervision for identifying driving-relevant evidence. Their grounded locations span the three-camera images and exhibit different spatial distributions across element classes, enabling models to associate heterogeneous visual evidence with its semantic role in driving (Fig.~\ref{fig:elements}). The free-text annotations remain compact, with mean lengths of \textbf{34} tokens for element-level implications and \textbf{32} tokens for rationales, supporting concise inference from individual element effects to integrated driving decisions. Finally, the action space comprises \textbf{4} longitudinal and \textbf{10} lateral actions, providing semantic planning supervision that connects grounded reasoning with motion planning (Fig.~\ref{fig:annotation_action}).

\section{Curriculum-Based Supervised Fine-Tuning}

\paragraph{Stage I: Perception and Grounding.}
Stage I focuses on learning visual perception and proceeds in two phases. In phase $a$, supervision is limited to scene context, traffic events, and binary decision-critical element presence, establishing coarse scene understanding before element-level reasoning. Phase $b$ introduces element identification, visual grounding, impact ranking, and attribute understanding. Throughout both phases, the complete VG-CoT target is retained, but locked fields are masked from the loss and contribute no gradient. This preserves a consistent output structure while allowing the model to progressively acquire the capabilities required for finer-grained perception.

\paragraph{Stage II: Reasoning and Planning.}
Stage II is initialized from the best Stage-I checkpoint and supervises the complete VG-CoT chain, including element-level implication, driving action rationale, action and trajectory generation. At this stage, the visual encoder is frozen to preserve the grounding capability learned in Stage I while the model learns higher-level reasoning and planning. We additionally apply trajectory-only supervision to a subset of training samples, where only trajectory tokens contribute to the loss, to preserve a direct perception-to-action pathway. We also increase the sampling probability of frames containing longitudinal action transitions and rare element states to improve the coverage of underrepresented cases.

\paragraph{Training Objective.}
For each target token $t$, we define a curriculum-aware weight $w_t = m_t^{(s)} W_{\phi(t)} \rho_t$, where $m_t^{(s)}$ is the stage-dependent supervision gate that determines which VG-CoT fields contribute to the loss at each curriculum. $W_{\phi(t)}$ is the base weight of the annotation field, and $\rho_t$ adjusts the contribution of informative or underrepresented targets. In particular, rare attribute values are upweighted, while trajectory waypoints that deviate more from constant-velocity motion receive greater emphasis.

The overall training objective combines a VG-CoT chain loss and a trajectory loss:
\begin{equation}
\mathcal{L}
=
\mathcal{L}_{\mathrm{chain}}
+
\lambda \mathcal{L}_{\mathrm{traj}},
\quad
\mathcal{L}_{\mathrm{chain}}
=
\frac{\sum_{t\in\mathcal{C}} w_t\ell_t}
{\sum_{t\in\mathcal{C}} w_t},
\quad
\mathcal{L}_{\mathrm{traj}}
=
\frac{\sum_{t\in\mathcal{J}} w_t\ell_t}
{\sum_{t\in\mathcal{J}} w_t}.
\end{equation}
Here, $\ell_t=-\log p_\theta(y_t\mid y_{<t},x)$ denotes the token-level cross-entropy loss. $\mathcal{C}$ and $\mathcal{J}$ denote the supervised VG-CoT chain tokens and trajectory tokens, respectively. $\mathcal{L}_{\mathrm{chain}}$ supervises the perception-to-action reasoning chain, while $\mathcal{L}_{\mathrm{traj}}$ optimizes future trajectory generation. The coefficient $\lambda$ controls the balance between the two objectives and is set to $1$ in all experiments. We normalize the two terms independently so that their relative weighting is not determined by differences in token sequence length. Stage I optimizes only $\mathcal{L}_{\mathrm{chain}}$, whereas Stage II jointly optimizes both objectives.

\section{Experiments}

\subsection{Models}
We evaluate eight representative models to assess the effectiveness of AnchorReasoning across diverse model foundations. \textbf{(i) General-purpose VLMs.} Qwen2.5-VL-7B \citep{bai2025qwen25vltechnicalreport} and Qwen3-VL-8B \citep{bai2025qwen3} serve as general-purpose baselines.\textbf{(ii) Embodied-AI foundation models.} Cosmos-Reason2-2B and Cosmos-Reason2-8B \citep{nvidia2025cosmosreason2} provide embodied-reasoning baselines without driving-specific adaptation. \textbf{(iii) AV-specific VLMs.} Alpamayo-R1-10B and Alpamayo-1.5-10B \citep{wang2025alpamayo} are driving-specialized models capable of joint reasoning and trajectory prediction. AutoVLA-3B \citep{zhou2026autovla} and Impromptu-VLA-7B \citep{chi2026impromptu} are included as complementary open-source driving baselines. Detailed training and implementation settings are provided in Appendix~\ref{app:hardware}, ~\ref{app:training_config}.

\subsection{Evaluation Metrics}
\paragraph{Standard Evaluation Metrics.}
We use standard classification, trajectory, and efficiency metrics to evaluate scene understanding, element attributes, trajectory prediction, and inference cost. Context Acc. measures the average accuracy of scene-context predictions, while Traffic Event F1 evaluates the detection of traffic events.  Presence Acc. measures whether the model correctly predicts the presence of decision-critical elements in a frame. For matched elements, Element State Recall, Element Intention Recall, and Sign-Content Recall evaluate the corresponding semantic attributes, while R/G Conf. measures the red--green traffic-light confusion rate. For trajectory prediction, we report Average Displacement Error (ADE) and Final Displacement Error (FDE). We additionally report the official WOD-E2E RFS~\citep{xu2026wod}. Detailed definitions and implementation protocols are provided in Appendix~\ref{app:evaluation_metrics}.

\label{sec:task_specific_metrics}
\paragraph{Decision-Critical Element Grounding.}
Because the model predicts a point rather than a bounding box, we match predicted elements to ground-truth elements using a one-to-one, scale-adaptive point-to-mask criterion. For each ground-truth element, the point-to-mask distance is zero when the predicted point falls inside its segmentation mask and otherwise equals the minimum Euclidean distance to the mask. A prediction is considered matched when this distance falls within a scale-adaptive assignment threshold determined by the ground-truth element size.

We report Element Recall $R_{\mathrm{det}}=|\mathcal{M}|/|\mathcal{G}|$, where $\mathcal{G}$ is the set of ground-truth decision-critical elements and $\mathcal{M}$ is the subset successfully matched to model predictions. We also report Rank-1 Recall to evaluate whether the most influential element is retrieved, Element Type Acc. to assess type prediction among matched elements, and Point-in-Mask Rate to measure localization precision:
\begin{equation}
R_{\mathrm{in}}
=
\frac{1}{|\mathcal{M}|}
\sum_{g\in\mathcal{M}}
\mathbf{1}\!\left[x_{\pi(g)}\in M_g\right],
\end{equation}

where $\pi(g)$ denotes the prediction matched to ground-truth element $g$, $x_{\pi(g)}$ is its predicted image point, $M_g$ is the segmentation mask of $g$, and $\mathbf{1}[\cdot]$ is the indicator function. Detailed matching rules and robustness analyses are provided in Appendix~\ref{app:evaluation_metrics}.

\paragraph{Element Implication and Driving Action Rationale.}
We use GPT-5.5 to evaluate element-level implications and frame-level rationales. Element Implication measures whether the model correctly describes each decision-critical element and its effect on ego driving, while Driving Action Rationale measures whether it correctly consolidates multiple factors into the final driving decision. Details are provided in Appendix~\ref{app:evaluation_metrics}.

\begin{table}[htbp]
\centering

\caption{
\textbf{Scene-level semantics and decision-critical element.}
}
\label{tab:sft_understanding}

\scriptsize
\setlength{\tabcolsep}{2.2pt}
\renewcommand{\arraystretch}{1.08}

\resizebox{\linewidth}{!}{%
\begin{tabular}{lccccccccccc}
\toprule
&
\multicolumn{2}{c}{\textbf{Scene-Level Semantics}}
&
\multicolumn{9}{c}{\textbf{Decision-Critical Elements}}
\\
\cmidrule(lr){2-3}
\cmidrule(lr){4-12}

\textbf{Model}
&
\makecell{\textbf{Context}\\\textbf{Acc.} $\uparrow$}
&
\makecell{\textbf{Traffic Event}\\\textbf{F1} $\uparrow$}
&
\makecell{\textbf{Presence}\\\textbf{Acc.} $\uparrow$}
&
\makecell{\textbf{Element}\\\textbf{Recall} $\uparrow$}
&
\makecell{\textbf{Rank-1}\\\textbf{Recall} $\uparrow$}
&
\makecell{\textbf{Element Type}\\\textbf{Acc.} $\uparrow$}
&
\makecell{\textbf{Point-in-Mask}\\\textbf{Rate} $\uparrow$}
&
\makecell{\textbf{Element State}\\\textbf{Recall} $\uparrow$}
&
\makecell{\textbf{Element Intention}\\\textbf{Recall} $\uparrow$}
&
\makecell{\textbf{Traffic Control}\\\textbf{-Content Recall} $\uparrow$}
&
\makecell{\textbf{R/G Conf.}\\$\downarrow$}
\\
\midrule

Qwen2.5-VL-7B
& -- / 0.83
& -- / 0.72
& 0.80 / 0.87
& 0.43 / 0.74
& 0.38 / 0.79
& 0.89 / 0.96
& 0.57 / 0.85
& 0.25 / 0.46
& 0.22 / 0.46
& -- / 0.95
& 0.042 / 0.029
\\

Qwen3-VL-8B
& -- / 0.82
& -- / 0.71
& 0.64 / 0.88
& 0.10 / 0.79
& 0.10 / 0.83
& 0.63 / 0.98
& 0.18 / 0.91
& 0.13 / 0.47
& 0.00 / 0.47
& -- / 0.95
& 0.037 / 0.026
\\

Cosmos-Reason2-2B
& -- / 0.83
& -- / 0.75
& 0.81 / 0.88
& 0.10 / 0.78
& 0.10 / 0.81
& 0.71 / 0.95
& 0.31 / 0.86
& 0.15 / 0.46
& 0.02 / 0.46
& -- / 0.98
& 0.029 / 0.025
\\

Cosmos-Reason2-8B
& -- / 0.82
& -- / 0.71
& 0.67 / 0.89
& 0.17 / 0.78
& 0.17 / 0.82
& 0.62 / 0.97
& 0.63 / 0.85
& 0.01 / 0.48
& 0.00 / 0.48
& -- / 0.95
& 0.034 / 0.013
\\

Alpamayo-R1-10B
& -- / 0.82
& -- / 0.74
& 0.83 / 0.89
& 0.49$^\dagger$ / 0.62
& 0.67 / 0.69
& -- / 0.90
& -- / 0.40
& -- / 0.45
& -- / 0.42
& -- / 0.92
& -- / 0.044
\\

Alpamayo-1.5-10B
& -- / 0.82
& -- / 0.70
& 0.84 / 0.91
& 0.46$^\dagger$ / 0.79
& 0.66 / 0.82
& -- / 0.98
& -- / 0.92
& -- / 0.71
& -- / 0.58
& -- / 0.97
& -- / 0.028
\\

AutoVLA-3B
& -- / 0.82
& -- / 0.72
& -- / 0.86
& -- / 0.79
& -- / 0.81
& -- / 0.95
& -- / 0.80
& -- / 0.46
& -- / 0.47
& -- / 0.97
& -- / 0.014
\\

Impromptu-VLA-7B
& -- / 0.83
& -- / 0.71
& -- / 0.86
& -- / 0.67
& -- / 0.76
& -- / 0.96
& -- / 0.85
& -- / 0.47
& -- / 0.48
& -- / 0.96
& -- / 0.028
\\

\bottomrule
\end{tabular}%
}

\vspace{2pt}
\begin{minipage}{\linewidth}
\scriptsize
\textit{Note.}
Values are reported as Base / VG-CoT.
``--'' denotes an unavailable or non-evaluable capability.
$\dagger$ indicates type-only matching because grounding ability is unavailable. $\uparrow$/$\downarrow$: higher/lower is better.
\end{minipage}
\vspace{-0.8em}
\end{table}

\subsection{Results}
To evaluate the effectiveness of VG-CoT supervision, we compare Base and VG-CoT models in scene-level semantics and decision-critical element understanding, reasoning and trajectory prediction, inference latency, and visual-grounding ablation. Base denotes the domain-adapted model trained on the original WOD-E2E data, while VG-CoT denotes the same model trained with VG-CoT supervision from AnchorReasoning. Training details are provided in Appendix~\ref{app:training_config}.

\begin{wrapfigure}{r}{0.6\textwidth}
    \vspace{-0.8em}
    \centering
    \includegraphics[width=\linewidth]{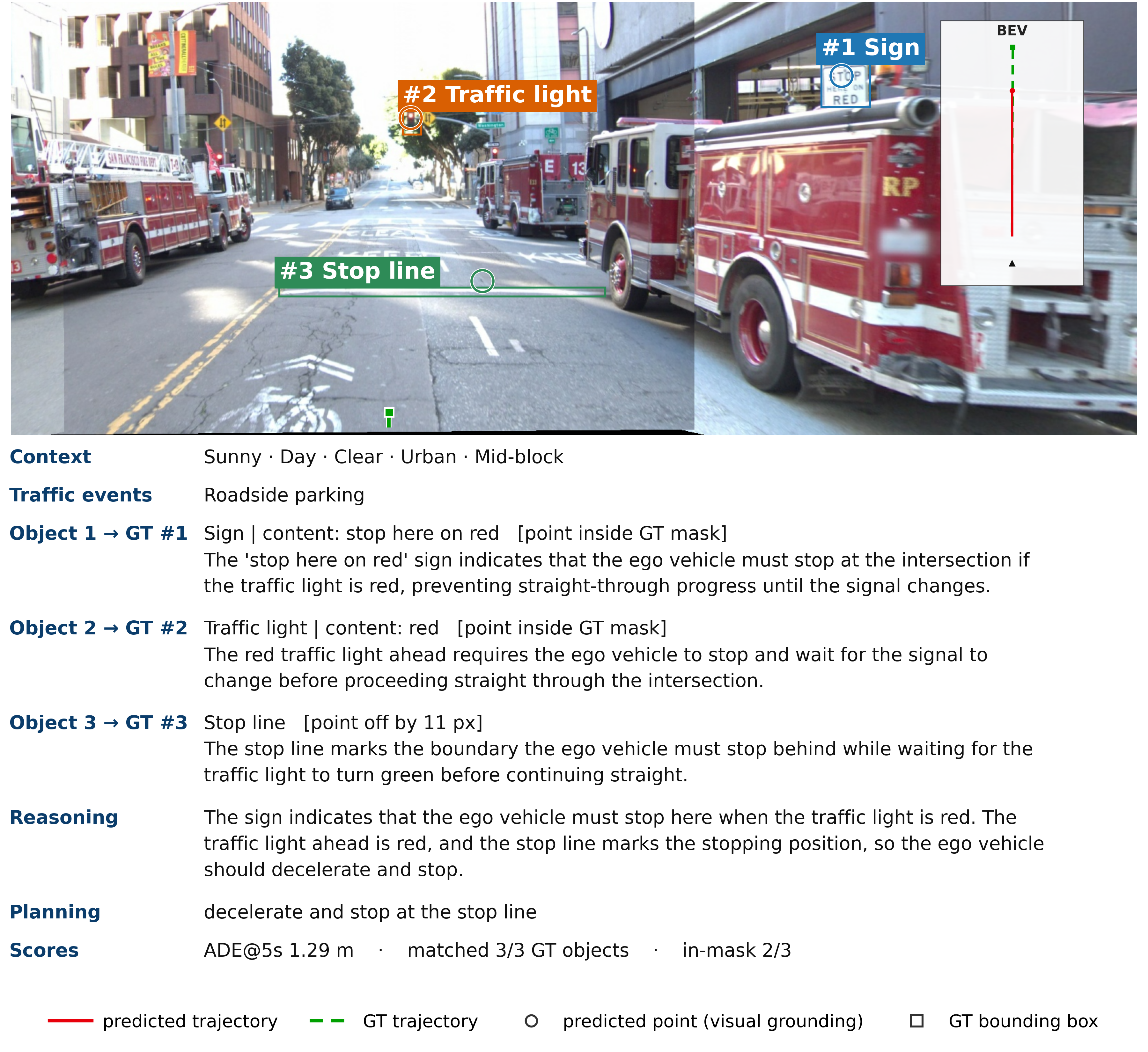}
    \caption{Example VG-CoT output of Alpamayo-1.5-10B.}
    \label{fig:reason-example}
    \vspace{-1.2em}
\end{wrapfigure}

\paragraph{Scene-Level Semantics and Decision-Critical Elements.}
Table~\ref{tab:sft_understanding} shows that VG-CoT supervision enables scene- and element-level predictions that are unavailable in several Base models, while improving the ability to identify and understand decision-critical elements. More importantly, the supervision enables models not only to state decision-critical elements in text but also to localize them in the image (Fig.~\ref{fig:reason-example}). For the general-purpose Qwen models and the Cosmos embodied-AI foundation models, the Base models can identify decision-critical elements and their types, but tend to retrieve only a limited number of such elements. For example, Qwen2.5-VL achieves 0.89 Element Type Acc. but only 0.43 Element Recall, while Cosmos-Reason2-2B achieves 0.71 type accuracy but only 0.10 Element Recall. After training with VG-CoT supervision, both Element Recall and Rank-1 Recall increase across these models. Moreover, Rank-1 Recall is higher than overall Element Recall for every trained backbone, indicating that the supervision more reliably captures the elements with the strongest influence on the driving decision. The AV-specific models exhibit different behavior. The Base Alpamayo models already show stronger decision-critical element retrieval than most general-purpose models, but lack explicit grounding and several structured attributes in their native outputs. VG-CoT supervision adds these capabilities; Alpamayo-1.5 reaches 0.92 Point-in-Mask Rate while maintaining strong Element Recall and Rank-1 Recall. In contrast, Alpamayo-R1 attains reasonable Rank-1 Recall but a much lower Point-in-Mask Rate of 0.40, showing that identifying an important element does not necessarily imply precise visual localization. Across all model families, Element Type Acc. and Traffic Control-Content Recall are consistently higher than Element State and Intention Recall, suggesting that dynamic behavior understanding remains more difficult than recognizing element identity and traffic-control semantics under single-frame visual input.

\begin{table}[htbp]
\centering

\caption{
\textbf{Reasoning, trajectory prediction, and inference efficiency.}}
\label{tab:sft_reasoning_trajectory_efficiency}

\scriptsize
\setlength{\tabcolsep}{1.8pt}
\renewcommand{\arraystretch}{1.08}

\resizebox{\linewidth}{!}{%
\begin{tabular}{lcccccccccc}
\toprule
&
\multicolumn{2}{c}{\textbf{Reasoning}}
&
\multicolumn{6}{c}{\textbf{Trajectory Prediction}}
&
\multicolumn{2}{c}{\textbf{Inference Efficiency}}
\\
\cmidrule(lr){2-3}
\cmidrule(lr){4-9}
\cmidrule(lr){10-11}

\textbf{Model}
&
\makecell{\textbf{Element}\\\textbf{Implication} $\uparrow$}
&
\makecell{\textbf{Driving Action}\\\textbf{Rationale} $\uparrow$}
&
\makecell{\textbf{ADE@1s}\\$\downarrow$}
&
\makecell{\textbf{ADE@3s}\\$\downarrow$}
&
\makecell{\textbf{ADE@5s}\\$\downarrow$}
&
\makecell{\textbf{FDE@5s}\\$\downarrow$}
&
\makecell{\textbf{RFS}\\\textbf{Frame} $\uparrow$}
&
\makecell{\textbf{RFS}\\\textbf{Cluster} $\uparrow$}
&
\makecell{\textbf{Avg.}\\\textbf{Tokens}}
&
\makecell{\textbf{Latency}\\\textbf{(s/frame)} $\downarrow$}
\\
\midrule

Qwen2.5-VL-7B
& 0.05 / 0.78
& 0.47 / 0.69
& 4.39 / \textbf{0.25}
& 8.81 / \textbf{1.48}
& 12.99 / \textbf{3.35}
& 24.15 / \textbf{8.23}
& 5.24 / \textbf{7.33}
& 5.27 / \textbf{7.39}
& 378 / 365
& 4.76 / 4.62
\\

Qwen3-VL-8B
& 0.54 / 0.79
& 0.36 / 0.72
& 3.72 / \textbf{0.22}
& 8.44 / \textbf{1.34}
& 12.95 / \textbf{3.04}
& 23.95 / \textbf{7.54}
& 5.37 / \textbf{7.54}
& 5.38 / \textbf{7.56}
& 384 / 398
& 4.77 / 4.90
\\

Cosmos-Reason2-2B
& 0.07 / 0.75
& 0.05 / 0.71
& 16.36 / \textbf{0.22}
& 24.03 / \textbf{1.35}
& 27.88 / \textbf{3.06}
& 36.27 / \textbf{7.58}
& 5.21 / \textbf{7.56}
& 5.20 / \textbf{7.59}
& 617 / 405
& 7.68 / 5.07
\\

Cosmos-Reason2-8B
& 0.35 / 0.80
& 0.22 / 0.72
& 5.84 / \textbf{0.22}
& 11.36 / \textbf{1.34}
& 16.23 / \textbf{3.02}
& 28.11 / \textbf{7.43}
& 5.28 / \textbf{7.56}
& 5.32 / \textbf{7.59}
& 529 / 389
& 9.12 / 6.74
\\

Alpamayo-R1-10B
& 0.61 / 0.70
& 0.60 / 0.61
& 0.24 / \textbf{0.23}
& 1.46 / \textbf{1.38}
& 3.37 / \textbf{2.97}
& 8.45 / \textbf{7.40}
& 7.05 / \textbf{7.66}
& 7.06 / \textbf{7.69}
& -- / 403
& -- / 7.00
\\

Alpamayo-1.5-10B
& 0.69 / 0.84
& 0.61 / 0.80
& 0.26 / \textbf{0.23}
& 1.55 / \textbf{1.34}
& 3.49 / \textbf{2.94}
& 8.56 / \textbf{7.06}
& 6.66 / \textbf{7.81}
& 6.69 / \textbf{7.83}
& -- / 359
& -- / 6.24
\\

AutoVLA-3B
& 0.48 / 0.76
& 0.32 / 0.69
& 0.62 / \textbf{0.30}
& 2.49 / \textbf{1.64}
& 5.01 / \textbf{3.60}
& 11.49 / \textbf{8.69}
& 6.18 / \textbf{7.17}
& 6.10 / \textbf{7.24}
& 341 / 373
& 5.68 / 6.21
\\

Impromptu-VLA-7B
& 0.53 / 0.75
& 0.54 / 0.70
& 0.23 / \textbf{0.22}
& 2.24 / \textbf{1.40}
& 5.96 / \textbf{3.17}
& 15.68 / \textbf{7.84}
& 6.31 / \textbf{7.45}
& 6.32 / \textbf{7.47}
& 144 / 352
& 1.91 / 4.46
\\

\midrule
\textbf{Average Improvement}
& +0.36
& +0.31
& -3.72
& -6.14
& -7.84
& -11.86
& +1.60
& +1.63
& -18.50
& -0.32
\\

\bottomrule
\end{tabular}%
}

\vspace{2pt}

\begin{minipage}{\linewidth}
\scriptsize
\textit{Note.}
Values are reported as Base / VG-CoT. Alpamayo models use ``-'' for efficiency metrics because their trajectories are generated by a separate diffusion module and are not directly comparable.
\end{minipage}
\vspace{-0.8em}
\end{table}

\paragraph{Reasoning, Trajectory Prediction, and Inference Efficiency.}

Table~\ref{tab:sft_reasoning_trajectory_efficiency} shows that VG-CoT supervision improves both Element Implication and Driving Action Rationale across all model families. Element Implication consistently scores higher than Driving Action Rationale, indicating that understanding the effect of an individual decision-critical element is easier than consolidating multiple factors into a final driving decision.

\begin{wraptable}{r}{0.56\textwidth}
\vspace{-0.8em}
\centering
\caption{Ablation results on trajectory prediction.}
\label{tab:ablation_traj}

\scriptsize
\setlength{\tabcolsep}{2.0pt}
\renewcommand{\arraystretch}{1.0}

\resizebox{\linewidth}{!}{%
\begin{tabular}{lcccccc}
\toprule
\textbf{Model}
& \makecell{\textbf{ADE}\\\textbf{@1s} $\downarrow$}
& \makecell{\textbf{ADE}\\\textbf{@3s} $\downarrow$}
& \makecell{\textbf{ADE}\\\textbf{@5s} $\downarrow$}
& \makecell{\textbf{FDE}\\\textbf{@5s} $\downarrow$}
& \makecell{\textbf{RFS}\\\textbf{Frame} $\uparrow$}
& \makecell{\textbf{RFS}\\\textbf{Cluster} $\uparrow$}
\\
\midrule

Alpamayo-1.5
& 0.26 / \textbf{0.23}
& 1.47 / \textbf{1.34}
& 3.26 / \textbf{2.94}
& 7.97 / \textbf{7.06}
& 7.10 / \textbf{7.81}
& 7.13 / \textbf{7.83}
\\

Qwen3-VL
& 0.32 / \textbf{0.22}
& 1.48 / \textbf{1.34}
& 3.21 / \textbf{3.04}
& 7.87 / \textbf{7.54}
& 7.44 / \textbf{7.54}
& 7.45 / \textbf{7.56}
\\

Cosmos-R2-2B
& 0.33 / \textbf{0.22}
& 1.53 / \textbf{1.35}
& 3.32 / \textbf{3.06}
& 8.00 / \textbf{7.58}
& 7.35 / \textbf{7.56}
& 7.39 / \textbf{7.59}
\\

\bottomrule
\end{tabular}%
}

\vspace{1pt}
\begin{minipage}{\linewidth}
\tiny
\textit{Note.} Values are reported as Ablation / VG-CoT.
\end{minipage}

\vspace{-0.6em}
\end{wraptable}

These reasoning gains are accompanied by improved trajectory prediction: ADE/FDE decrease and both RFS metrics increase with VG-CoT supervision. The visual-grounding ablation in Table~\ref{tab:ablation_traj} further shows that VG-CoT consistently outperforms its non-grounded variant across all trajectory metrics for the three evaluated backbones, indicating that visually grounded CoT provides additional benefits for trajectory prediction. Despite the additional reasoning structure, VG-CoT does not require longer outputs. The average output length decreases by 18.5 tokens and latency by 0.32\,s/frame while trajectory accuracy improves, suggesting that VG-CoT can provide more compact reasoning without sacrificing planning performance.

\section{Limitations and Future Work}
Our annotations currently cover only the front-left, front, and front-right camera views, leaving side and rear views unannotated. Future work will extend AnchorReasoning to full-surround perception, potentially using trained models to assist annotation. We also lack explicit depth supervision. Since depth and depth change provide important cues for identifying decision-critical elements, future work will incorporate depth estimates to strengthen spatial understanding and reasoning.

\section{Conclusion}
We introduce AnchorReasoning, a long-tail autonomous-driving dataset for VLMs that provides visually grounded supervision over decision-critical elements, their attributes and implications, driving-action rationale, and action and trajectory planning. Its VG-CoT structure explicitly connects visual evidence with reasoning and planning, while the curriculum SFT strategy enables models to progressively acquire these capabilities. We further introduce an object-size-aware grounding metric for evaluating decision-critical element localization. Experiments across eight general-purpose, embodied-AI, and AV-specific backbones show that VG-CoT supervision improves grounded reasoning and trajectory prediction. Visual-grounding ablations further confirm that CoT with explicit visual grounding benefits trajectory prediction. Across models, VG-CoT reduces 5-s ADE and FDE by 7.84 and 11.86 on average, improves RFS Frame and Cluster by 1.66 and 1.70, and reduces inference latency by 0.32\,s/frame with fewer reasoning tokens. These results demonstrate the value of visually grounded, decision-focused supervision for enhancing VLM reasoning and planning in long-tail autonomous driving.

% ---------------------------------------------------------------------------
% ICLR 2027 end-of-paper statements (do not count toward the page limit).
% ---------------------------------------------------------------------------
\subsection*{AI use statement}
% REQUIRED by ICLR 2027. See the ICLR 2027 AI Policy for Authors; at most 1 page.
Generative AI tools were used to assist with dataset annotation and
data cleaning and reformatting. AI tools were also used for code writing
and modification and for language editing of the manuscript to improve
clarity and readability. All AI-assisted annotations, data processing
outputs, code, and text were reviewed and verified by the authors before
use. The authors take full responsibility for the final content, results,
and claims of this work.

\subsection*{Ethics statement}
This work is built upon the Waymo Open Dataset. We do not redistribute
the original Waymo data; only the annotations introduced in this
work are released. Users must obtain the corresponding source data
separately under the terms of the Waymo Open Dataset license.

% \subsection*{Reproducibility statement}
% % (Recommended.) Point to the appendix / supplementary material / code.
% Additional implementation and experimental details are provided in the
% appendix after the references. The code is included in the supplementary material. The dataset, model weights, and code will be publicly released upon
% acceptance.
% Acknowledgments must stay hidden in the anonymous submission.

\bibliography{refs}
\bibliographystyle{iclr2027_conference}

\appendix
\section{Dataset Details}
\label{app:dataset_details}

\subsection{Dataset Composition and WOD-E2E Splits}
\label{app:dataset_composition}

AnchorReasoning is constructed on top of the Waymo Open Dataset for End-to-End Driving (WOD-E2E) without modifying the original sensor data or dataset splits. We add semantic and reasoning annotations directly to the decoded WOD-E2E frames. Table~\ref{tab:dataset_split} summarizes the correspondence between our annotations and the decoded WOD-E2E splits.

\begin{table}[H]
\centering
\caption{Composition of AnchorReasoning with respect to our decoded copy of the WOD-E2E splits.}
\label{tab:dataset_split}
\small
\begin{tabular}{lrrr}
\toprule
\textbf{Split} &
\textbf{\# Decoded Segments} &
\textbf{\# Annotated Segments} &
\textbf{\# Annotated Frames} \\
\midrule
Train      & 2,037 & 2,037 & 415,663 \\
Validation & 479   & 455   & 456 \\
Test       & 1,505 & 0     & 0 \\
\midrule
\textbf{Total annotated} & -- & \textbf{2,492} & \textbf{416,119} \\
\bottomrule
\end{tabular}
\end{table}

For the training split, all 415,663 decoded frames from 2,037 segments are annotated. For validation, we annotate the 456 frames for which WOD-E2E provides rater-feedback trajectories. These frames come from 455 of the 479 decoded validation segments and are used for both model evaluation and Rater Feedback Score (RFS) computation. Each frame is associated with three rated candidate trajectories, resulting in 1,368 human-rated trajectories in total. The remaining decoded validation frames are not included in the annotation statistics. The decoded test split is retained for completeness but is not annotated. Overall, AnchorReasoning contains 416,119 annotated frames from 2,492 WOD-E2E segments and 395,379 decision-critical elements.

\subsection{Impact-Rank Distribution}
\label{app:impact_rank}
\begin{figure}[htbp]
    \centering
    \includegraphics[width=\linewidth]{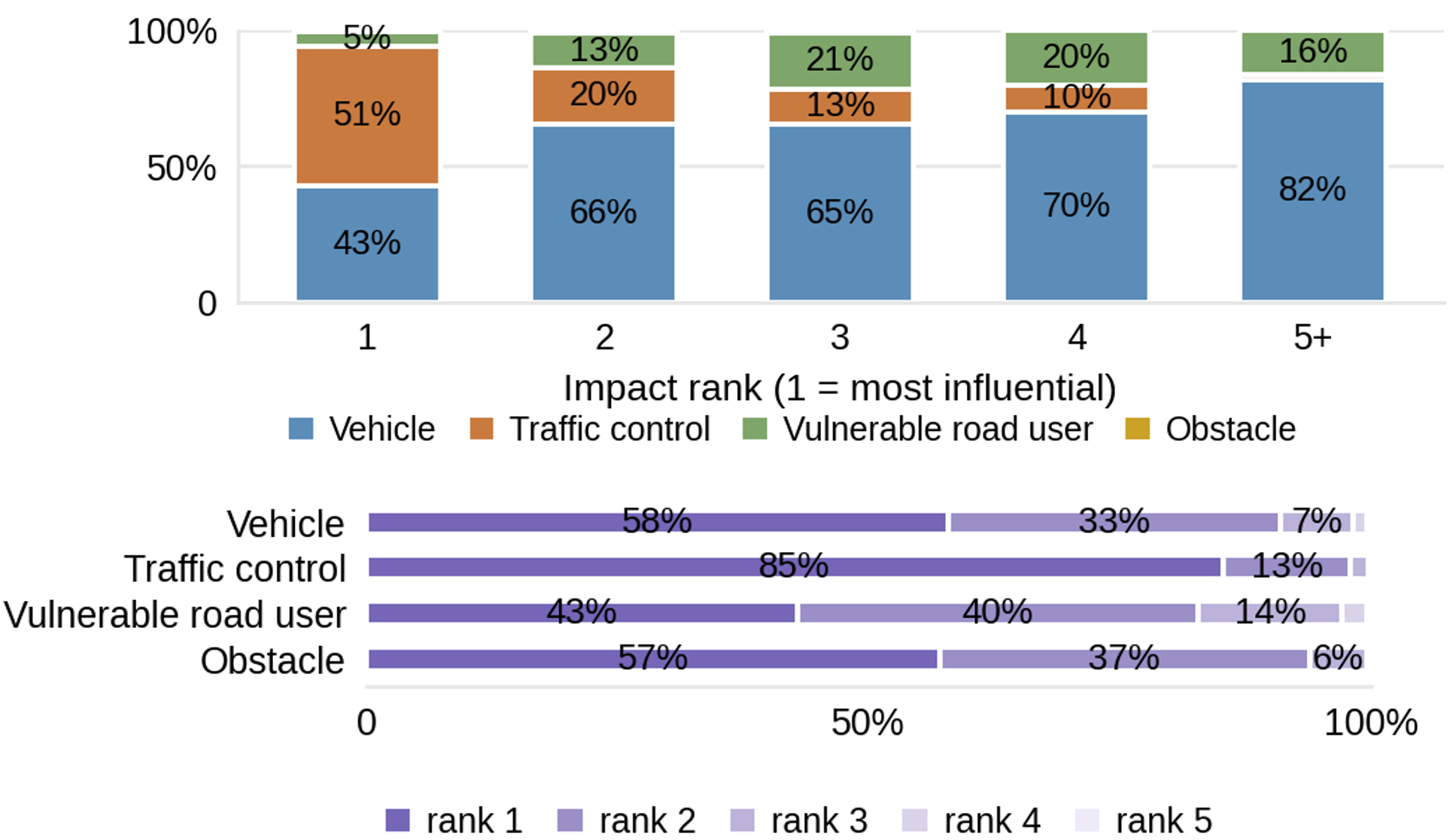}
    \caption{Distribution of Impact Ranks}
    \label{fig:class_rank}
\end{figure}

The distribution is strongly concentrated at the highest ranks: 67.9\% of the annotated elements are assigned rank 1, 25.3\% rank 2, 5.6\% rank 3, and 1.0\% rank 4, while only 0.2\% have rank 5 or higher. 93.2\% of all decision-critical elements fall within the first two ranks and 98.8\% within the first three, as shown in Fig.~\ref{fig:class_rank}. This distribution reflects the annotation principle of retaining only elements that materially influence the current driving decision rather than exhaustively labeling all visible objects.

\subsection{Action-Plan Distribution}
\label{app:action_distribution}
\begin{figure}[htbp]
    \centering
    \includegraphics[width=\linewidth]{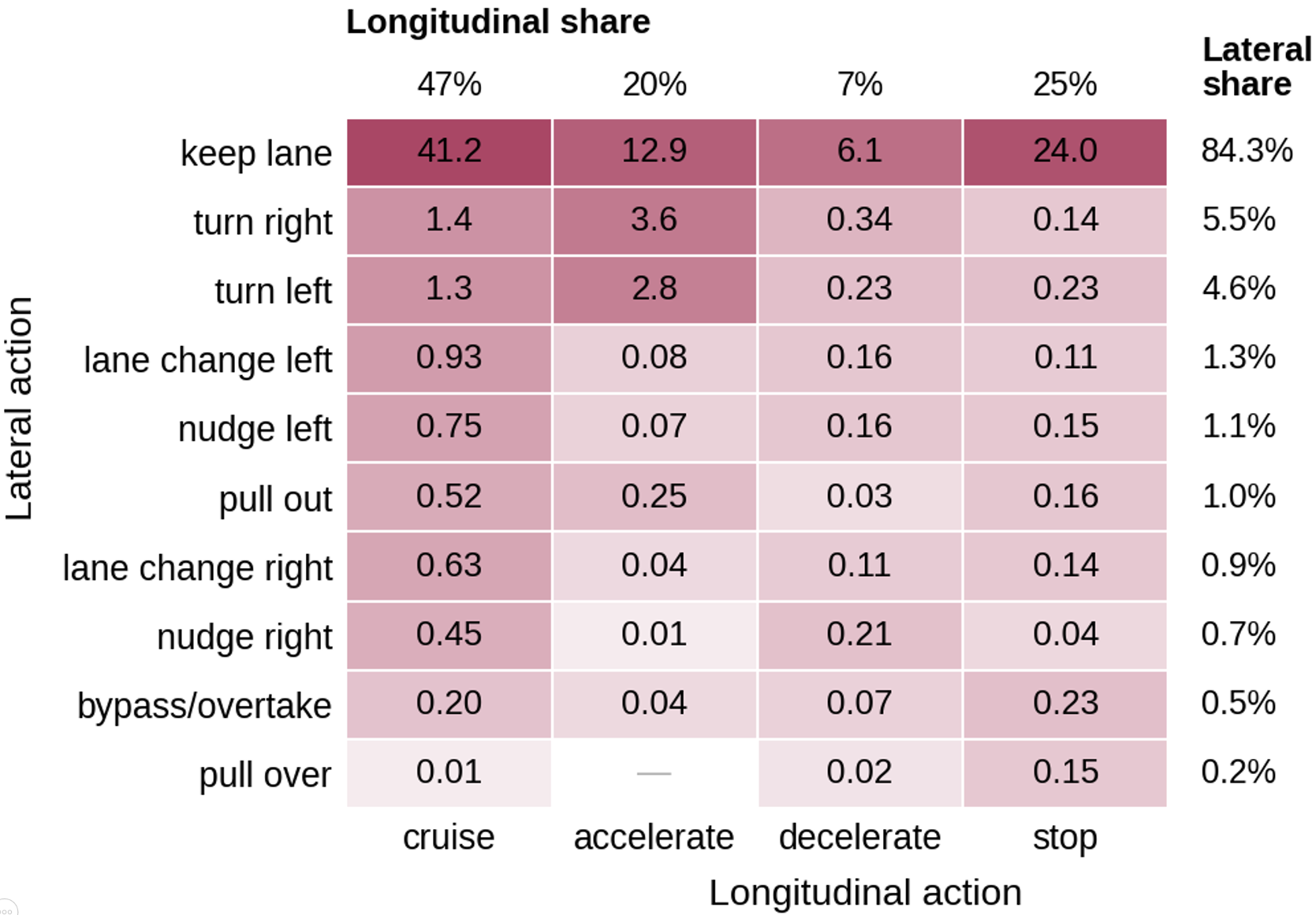}
    \caption{Joint distribution of the normalized lateral and longitudinal actions.}
    \label{fig:final_plan_distribution}
\end{figure}

Figure~\ref{fig:final_plan_distribution} shows the joint distribution of lateral and longitudinal actions. The distribution is naturally imbalanced, with routine lane-keeping action accounting for most frames and more complex lateral maneuvers forming a long tail. Importantly, lateral and longitudinal actions are not independent: the same lateral maneuver can be associated with substantially different longitudinal responses depending on the surrounding constraints and interaction state. The dataset therefore provides supervision not only for recognizing individual actions, but also for learning their joint behavioral structure.

\subsection{Annotation Examples}
\label{app:annotation_examples}
\begin{figure}[htbp]
    \centering
    \includegraphics[width=\linewidth]{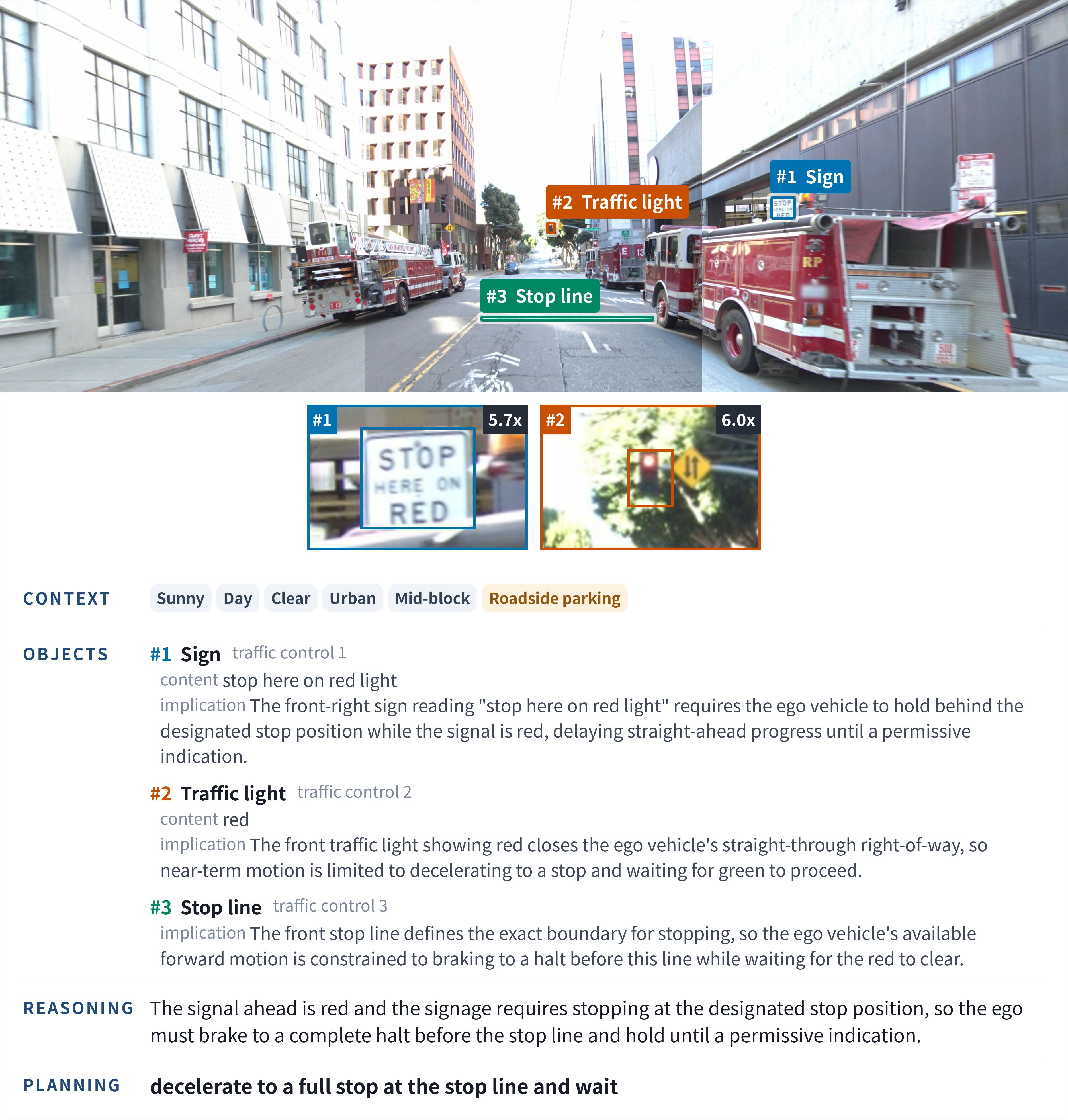}
    \caption{Examples of Complete Visual and Textual Annotations.}
    \label{fig:text-samples}
\end{figure}

\begin{figure}[htbp]
    \centering
    \includegraphics[width=\linewidth]{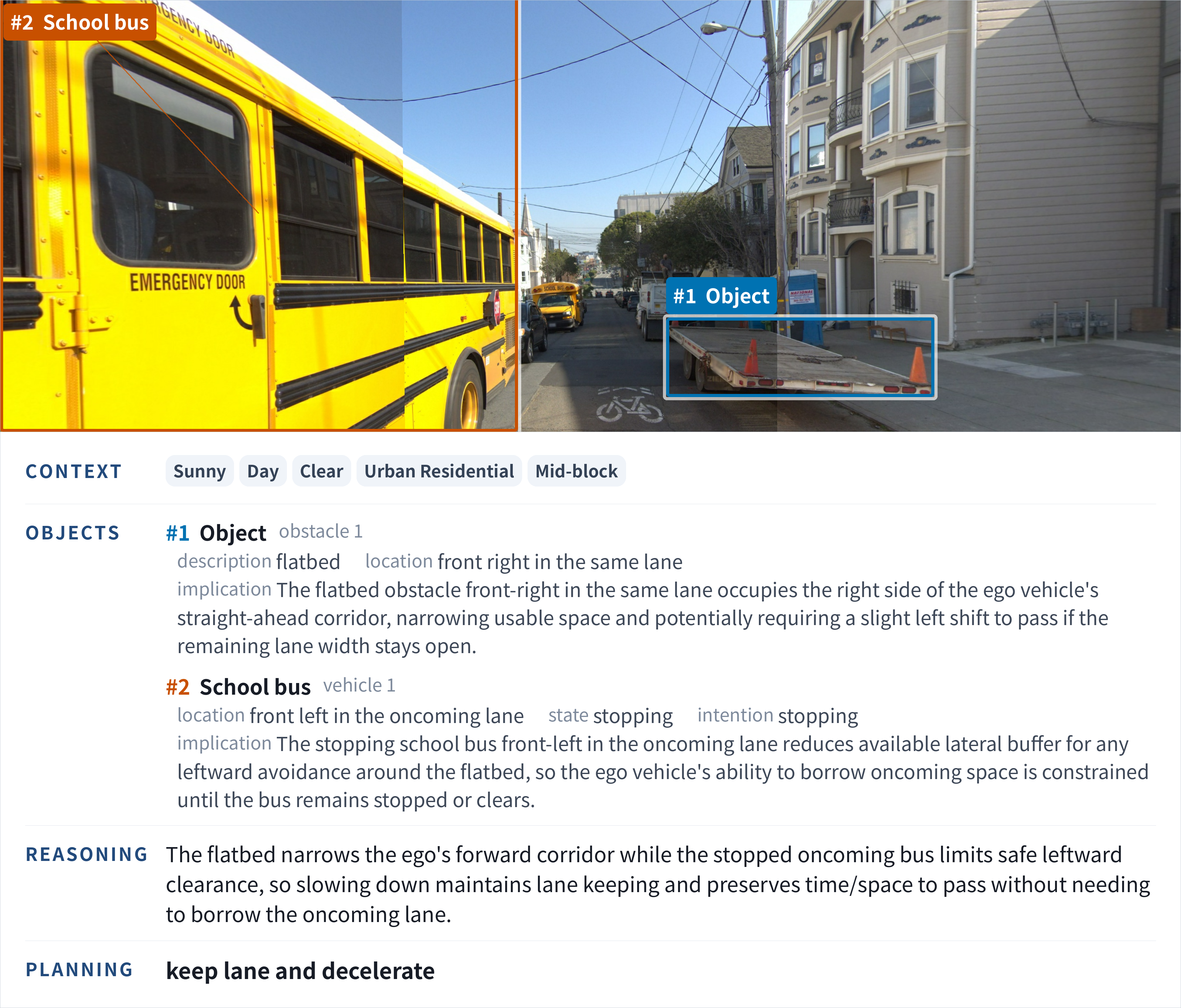}
    \caption{Examples of Complete Visual and Textual Annotations.}
    \label{fig:text-samples2}
\end{figure}

\begin{figure}[htbp]
    \centering
    \includegraphics[width=\linewidth]{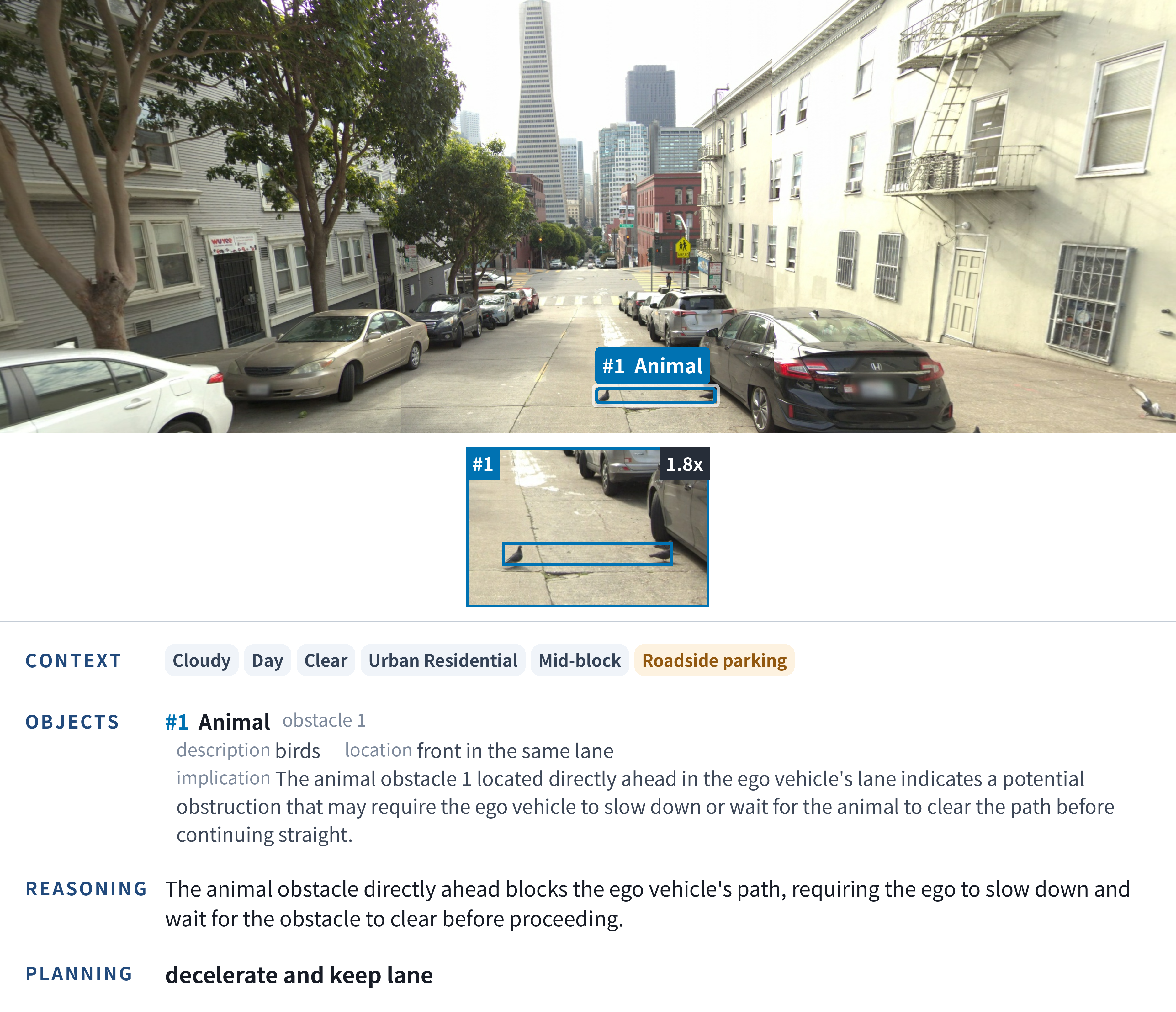}
    \caption{Examples of Complete Visual and Textual Annotations.}
    \label{fig:text-samples3}
\end{figure}

Figure~\ref{fig:text-samples}-~\ref{fig:text-samples3} present the representative examples of the AnchorReasoning annotation. The annotation progressively connects scene context, decision-critical element grounding and attributes, element-level driving implications, frame-level reasoning, and the final action plan.

\section{Automatic Annotation Tools}
\label{app:auto_annotation}

This section describes the automatic tools used to assist dataset annotation and evaluates their reliability before large-scale deployment. To assess annotation quality, we use a manually annotated subset of the dataset as reference and compare the outputs of each automatic tool against the corresponding human labels. The automatic tools are applied independently without access to these reference labels. Because human annotations may themselves contain ambiguity, the reported results measure human--automatic agreement rather than absolute annotation accuracy.

\subsection{Scene Context}
\label{app:context}

\paragraph{Method.}
We use Qwen3-VL-8B-Instruct to classify five scene-level attributes from the image: weather, time of day, visibility, scenario type, and road structure. The model receives only the image and a predefined label taxonomy, and returns a structured JSON output containing the predicted value, confidence, and brief visual evidence. To reduce frame-level fluctuations, predictions are further stabilized within each video. Weather and visibility are kept temporally consistent, missing scenario labels are propagated from adjacent frames, and isolated road-structure changes are detected and corrected. The complete prompt used for scene-context annotation is shown in Fig.~\ref{fig:context_prompt}.

\paragraph{Reliability Evaluation.}
We evaluate the context classifier on 7,450 manually annotated frames. For each context field, human--automatic agreement is computed as
\begin{equation}
A =
\frac{1}{N}
\sum_{i=1}^{N}
\mathbf{1}
\left[
\hat{y}_i = y_i^{\mathrm{human}}
\right],
\end{equation}
where $y_i^{\mathrm{human}}$ is the human annotation, $\hat{y}_i$ is the automatic prediction, and $\mathbf{1}[\cdot]$ denotes the indicator function.

\begin{table}[H]
\centering
\caption{Human--automatic agreement for scene-context annotations.}
\label{tab:context_quality}
\small
\begin{tabular}{lcc}
\toprule
\textbf{Field} & \textbf{\# Frames} & \textbf{Agreement} \\
\midrule
Time of day    & 7,450 & 100.00\% \\
Visibility     & 7,450 & 88.68\% \\
Weather        & 7,450 & 83.57\% \\
Road structure & 7,450 & 82.32\% \\
Scenario       & 7,450 & 76.51\% \\
\bottomrule
\end{tabular}
\end{table}

\subsection{Traffic-Light State}
\label{app:traffic_light}

\paragraph{Method.}
Traffic-light states are determined using a rule-based HSV classifier. For each annotated traffic light, pixels are extracted from its instance segmentation mask rather than the full bounding box to reduce background contamination. RGB pixels are converted to HSV space, and bright pixels with $V \geq 150$ are selected for analysis. If fewer than 10 bright pixels are available, all pixels within the instance mask are used instead. The adopted HSV thresholds are shown in Table~\ref{tab:hsv_thresholds}. A color is considered detected only when at least 10 pixels satisfy the corresponding thresholds and account for at least 0.5\% of the valid pixels.

\begin{table}[htbp]
\centering
\caption{HSV thresholds used for traffic-light state recognition.}
\label{tab:hsv_thresholds}
\small
\begin{tabular}{lccc}
\toprule
\textbf{State} & \textbf{Hue Range} & $\mathbf{S_{\min}}$ & $\mathbf{V_{\min}}$ \\
\midrule
Red    & $[0,25]\cup[160,180]$ & 50 & 80 \\
Yellow & $[20,50]$              & 25 & 120 \\
Green  & $[50,95]$              & 40 & 40 \\
\bottomrule
\end{tabular}
\end{table}

\paragraph{Reliability Evaluation.}
Evaluation is performed on three manually annotated partitions containing 11,318 matched traffic-light instances. The classifier achieves 97.81\% overall agreement. Red and green recognition is highly reliable, while yellow remains the primary source of error.

\begin{table}[t]
\centering
\caption{Traffic-light recognition results on the evaluation partitions.}
\label{tab:traffic_light_eval}
\small
\begin{tabular}{lcccc}
\toprule
\textbf{Partition} &
\textbf{\# Lights} &
\textbf{Agreement} &
\textbf{Red Recall} &
\textbf{Green Recall} \\
\midrule
p19 & 5,300 & 98.28\% & 0.9935 & 0.9938 \\
p20 & 4,869 & 97.00\% & 0.9949 & 0.9733 \\
p21 & 1,149 & 99.04\% & 0.9832 & 1.0000 \\
\midrule
All & 11,318 & 97.81\% & -- & -- \\
\bottomrule
\end{tabular}
\end{table}

The red--green confusion rate ranges from 0.40\% to 1.34\%. In contrast, yellow recall is 0.42 on p19 and 0.59 on p20, despite relatively high precision. This motivates the conservative labeling strategy used in dataset construction: automatic predictions are used to lbael red and green states. Human annotate the yellow labels.

We additionally evaluate three threshold configurations to examine the trade-off between yellow recall and false yellow detections. The results on p19 are summarized in Table~\ref{tab:traffic_light_ablation}.

\begin{table}[H]
\centering
\caption{Ablation of traffic-light color thresholds.}
\label{tab:traffic_light_ablation}
\small
\begin{tabular}{lccc}
\toprule
\textbf{Setting} &
\textbf{Overall Agreement} &
\textbf{Yellow Recall} &
\textbf{Yellow Precision} \\
\midrule
Adopted        & 98.28\% & 0.42 & 0.95 \\
Relaxed yellow & 73.47\% & 0.87 & 0.06 \\
Intermediate   & 97.02\% & 0.69 & 0.41 \\
\bottomrule
\end{tabular}
\end{table}

Relaxing the yellow threshold substantially improves yellow recall but introduces a large number of false yellow predictions from red lights. The adopted configuration therefore provides a more reliable overall trade-off.

\subsection{Object Location}
\label{app:location}

\paragraph{Method.}
Object bearing is labeled using the fixed geometry of the iamge. Five spatial regions are manually defined on a reference image and reused for all frames.

For each vehicle, vulnerable road user, or obstacle, we compute the overlap ratio between its bounding box $B$ and each spatial region $R_k$:
\begin{equation}
s_k =
\frac{\mathrm{Area}(B \cap R_k)}
     {\mathrm{Area}(B)}.
\end{equation}
The object is assigned to the region with the largest overlap. When the difference between the two highest overlap scores is no greater than 0.05, the bottom-center point of the bounding box is used as an additional point-in-polygon cue because it more closely approximates the object's contact position on the road.

Only the directional prefix of the location label is automatically labeled, while the road-topological relation annotation is labeled by human. For example, for the label \emph{front left in the oncoming lane}, only \emph{front left} is automatically labeled.

\paragraph{Reliability Evaluation.}
We compare the automatically determined directional prefix with the human annotation on 4,439 aligned objects. The two annotations agree on 4,369 instances, corresponding to an agreement rate of 98.42\%. Most disagreements occur between neighboring spatial regions, such as \emph{front} and \emph{front left}, rather than between opposite directions. This indicates that the remaining errors are primarily associated with region boundaries.

\subsection{Vehicle Type}
\label{app:vehicle_type}

\paragraph{Method.}
Vehicle types are classified using Qwen3-VL-8B-Instruct. Each vehicle is cropped from the image using its bounding box with a 20-pixel margin and classified into one of seven categories: \emph{Car}, \emph{Truck}, \emph{Emergency vehicle}, \emph{Bus}, \emph{Construction vehicle}, \emph{School bus}, and \emph{Other}. The prompt is shown in Fig.~\ref{fig:vehicle-type-prompt}

\paragraph{Reliability Evaluation.}
The classifier is evaluated on 3,720 manually labeled vehicles and agrees with the human annotations on 3,681 instances, yielding 98.95\% agreement. The remaining disagreements are concentrated around visually similar vehicle categories, particularly the boundary between cars, pickup trucks, and large van-like vehicles.

\subsection{Reasoning Chain Generation}
\label{app:reasoning_chain}

\paragraph{Method.}
The reasoning chain is generated in two stages. First, an element-level driving implication is generated for each decision-critical element. The inputs include the grounded iamge, scene context, ego state, navigation intent, and the structured attributes of the element. The implication describes how the element affects the ego vehicle's intended path, available motion, right-of-way, or local maneuver options. Each element is interpreted independently so that its causal effect is established before multiple factors are integrated. Second, the element-level implications are combined with their impact ranks, ego state, and navigation intent to generate a frame-level rationale and final action plan. The rationale integrates the dominant constraints and enabling factors, while the final action plan specifies the resulting short-horizon longitudinal and lateral action. The plan is constrained to the action vocabulary used in AnchorReasoning.

To balance annotation quality and computational cost, scenes are routed according to annotation complexity. Simple scenes, defined as those containing at most one decision-critical element or exactly two traffic-control elements, are processed using a locally deployed Qwen3.5-122B model, while more complex scenes are processed using GPT-5.2. All generated implications, rationales, and action plans are subsequently inspected and corrected through a dedicated annotation interface. Consecutive frames with identical annotations may reuse the previous output to improve temporal consistency and reduce unnecessary inference. The complete prompt used for driving implication annotation and rationale is shown in Fig.~\ref{fig:implication-prompt}, ~\ref{fig:reasoning-planning-prompt}.

\subsection{Action--Trajectory Consistency Check.}
\label{app:conssistency}
After annotation, we apply a rule-based consistency check to verify whether each annotated action plan agrees with the corresponding future trajectory. The action plan is mapped to one of four longitudinal categories,
\{\emph{accelerate}, \emph{decelerate}, \emph{cruise}, \emph{stop}\},
and, when explicitly specified, to one of three lateral categories,
\{\emph{left turn}, \emph{right turn}, \emph{straight}\}.

The future trajectory is converted to the same motion representation using kinematic and geometric rules. Let $v_0$ denote the current ego speed, $v_{\mathrm{end}}$ the mean speed over the final 1\,s of the future trajectory, and $v_{\max}$ the maximum future speed. The longitudinal trajectory category is defined as

\begin{equation}
\psi_{\mathrm{lon}}(\tau)=
\begin{cases}
\mathrm{stop},
& v_{\mathrm{end}} < 0.5,\\

\mathrm{accelerate},
& (v_0 < 0.5 \land v_{\max}\geq1.0)
\lor
(v_{\mathrm{end}}-v_0 > \max(1.0,0.2v_0)),\\

\mathrm{decelerate},
& v_0-v_{\mathrm{end}} > \max(1.0,0.2v_0)
\land v_{\mathrm{end}}\geq0.5,\\

\mathrm{cruise},
& \mathrm{otherwise}.
\end{cases}
\end{equation}

For lateral motion, the trajectory is classified as a left or right turn when the absolute heading change exceeds $25^\circ$, with the sign determining the turning direction; otherwise, it is classified as straight.

Consistency is checked using the compatibility rules rather than strict category equality. For example, an annotated \emph{decelerate} action is compatible with a trajectory that decelerates or stops, whereas an annotated \emph{stop} action is inconsistent with a trajectory that immediately continues at normal speed. A sample is considered consistent only when its longitudinal action and any explicitly specified lateral action are non-contradictory. Inconsistent samples are flagged for manual review and correction.

\section{Training and Evaluation Details}
\label{app:experimental_setup}

\subsection{Hardware and Implementation Details}
\label{app:hardware}

All training and evaluation experiments are conducted on NVIDIA B200 GPUs, with two GPUs used for training. Training uses distributed data parallelism with DeepSpeed ZeRO Stage~2 and bfloat16 precision. FlashAttention-2 is enabled for efficient attention computation, and gradient checkpointing is applied to reduce memory consumption. We use full-parameter fine-tuning for all trainable components without LoRA or other parameter-efficient adaptation methods.

\subsection{Training Configuration}
\label{app:training_config}
For both domain adaptation and our Supervised Fine-Tuning(SFT), each sample uses the same input representation: a single panorama stitched from three forward-facing cameras, the ego-vehicle motion history, and the high-level navigation intent. The motion history contains 16 waypoints over the preceding 4\,s and is provided as textual input. No temporal image sequence, LiDAR, HD map, or auxiliary prediction head is used.

\paragraph{Domain Adaptation.}
For the domain-adapted baselines, training is performed on the original WOD-E2E data without using the additional AnchorReasoning annotations. We preserve each model's original training objective, output format, and training configuration as much as possible. The input also follows each model's original formulation, with the only modification being the visual input, which is replaced by a single-frame panorama stitched from three forward-facing cameras. The target output follows each model's original formulation, such as its native reasoning, action, or trajectory representation, rather than the  visually grounded chain-of-thought (VG-CoT) annotation format.

\paragraph{Supervised Fine-Tuning.}
Our SFT uses the same input representation but supervises the model with the structured VG-CoT output. Stage~I focuses on scene-level semantics and decision-critical element understanding, including traffic events, element presence, element identification, visual grounding, impact rank, and element attributes. Stage~II is initialized from the best Stage-I checkpoint and extends the output to the complete VG-CoT chain, including ego-state understanding, element-level implications, driving-action rationale, action planning, motion, and future trajectory.

Table~\ref{tab:training_hyperparameters} summarizes the optimization settings used for SFT across all backbones. Stage~I starts from the original pretrained or released backbone, whereas Stage~II is initialized from the best Stage-I checkpoint. The same optimization settings are used across backbones unless otherwise specified.

\begin{table}[H]
\centering
\caption{\textbf{Training hyperparameters for Stage I and Stage II.}}
\label{tab:training_hyperparameters}

\small
\setlength{\tabcolsep}{4.5pt}
\renewcommand{\arraystretch}{1.08}

\resizebox{\linewidth}{!}{%
\begin{tabular}{lcc@{\hspace{14pt}}lcc}
\toprule
\textbf{Hyperparameter}
& \textbf{Stage I}
& \textbf{Stage II}
& \textbf{Hyperparameter}
& \textbf{Stage I}
& \textbf{Stage II}
\\
\midrule

Epochs
& 2 & 2
&
Optimizer
& AdamW & AdamW
\\

Per-device batch size
& 1 & 1
&
Adam $\beta_1,\beta_2$
& 0.9, 0.999 & 0.9, 0.999
\\

Effective batch size
& 64 & 64
&
Adam $\epsilon$
& $10^{-8}$ & $10^{-8}$
\\

Learning rate
& $1\times10^{-5}$ & $5\times10^{-6}$
&
Weight decay
& 0 & 0
\\

LR scheduler
& Cosine & Cosine
&
Max. gradient norm
& 1.0 & 1.0
\\

Warmup ratio
& 0.03 & 0.03
&
Precision
& bf16 & bf16
\\

Visual encoder
& Trainable & Frozen
&
DeepSpeed
& ZeRO-2 & ZeRO-2
\\

Visual LR multiplier
& 0.1 & --
&
Random seed
& 42 & 42
\\

\bottomrule
\end{tabular}%
}
\end{table}

\subsection{Evaluation Metrics}
\label{app:evaluation_metrics}

\paragraph{Scene and Element-Level Semantics.}
We evaluate scene understanding using Context Acc. and Traffic Event F1. Context Acc. is the macro-average accuracy over five scene-context dimensions: weather, daytime, visibility, scenario type, and road structure. Traffic Event F1 measures set-level agreement between predicted and annotated traffic events. Presence Acc. is the frame-level binary accuracy of whether at least one decision-critical element is present. Element State Recall and Element Intention Recall are macro-averaged recall scores over normalized behavior categories for matched vehicles and vulnerable road users. Sign-Content Recall evaluates the semantic content of matched traffic signs using macro-averaged recall over the closed-set control labels. R/G Conf. measures the fraction of red and green traffic lights that are confused with each other; lower values are better.

\paragraph{Decision-Critical Element Grounding.}
The model emits one grounding point for each predicted decision-critical element. Because no bounding box is predicted, conventional IoU-based matching is not applicable. We therefore use a point-to-region distance defined with respect to the ground-truth segmentation mask. For a predicted point $x$ and ground-truth element $g$,
\begin{equation}
d(x,g)=
\begin{cases}
0, & x\in M_g,\\
\min\limits_{p\in M_g}\|x-p\|_2, & \text{otherwise},
\end{cases}
\end{equation}
where $M_g$ denotes the segmentation mask of $g$. If a mask is unavailable, we fall back to the distance to the ground-truth bounding-box boundary rather than its center.

Predicted and ground-truth elements are matched one-to-one using a scale-adaptive assignment threshold
\begin{equation}
\tau_{\mathrm{assign}}(g)
=
\operatorname{clip}
\left(
1.5\,d_g,\,
60,\,
300
\right),
\end{equation}

where $d_g$ is the diagonal length of the ground-truth bounding box in pixels. Ground-truth elements are processed in increasing numerical impact rank, with rank~1 matched first, and each element claims the nearest unmatched prediction within $\tau_{\mathrm{assign}}$. Matching is type-agnostic so that localization and semantic classification are evaluated independently. Unmatched ground-truth elements are counted as misses, while unmatched predictions are treated as false positives.

Element Recall is defined as
\begin{equation}
R_{\mathrm{det}}
=
\frac{|\mathcal{M}|}{|\mathcal{G}|},
\end{equation}

where $\mathcal{G}$ is the set of ground-truth decision-critical elements and $\mathcal{M}$ is the subset successfully matched to predictions. Rank-1 Recall applies the same criterion only to elements with impact rank~1. Element Type Acc. measures semantic classification accuracy over the matched elements.

Localization precision is evaluated using Point-in-Mask Rate,
\begin{equation}
R_{\mathrm{in}}
=
\frac{1}{|\mathcal{M}|}
\sum_{g\in\mathcal{M}}
\mathbf{1}\!\left[x_{\pi(g)}\in M_g\right],
\end{equation}

where $\pi(g)$ denotes the prediction matched to $g$ and $x_{\pi(g)}$ is its predicted image point. Unlike bounding-box containment, this metric requires the predicted point to fall on the actual segmented object rather than on background pixels inside its bounding box.

\paragraph{Reasoning.}
Element-level implications and frame-level rationales are open-ended textual outputs for which lexical-overlap metrics are poorly aligned with semantic correctness. We therefore evaluate them using GPT-5.5 as a judge.

For Element Implication, the judge assesses whether the prediction correctly characterizes the corresponding decision-critical element, including its driving-relevant state or content, and whether it correctly describes the element's influence on ego driving. For Driving Action Rationale, the judge assesses whether the prediction appropriately integrates the relevant scene factors, identifies the dominant factor shaping the driving decision, and correctly explains how that factor influences ego action.

For each prediction, GPT-5.5 returns two binary judgments, $s_i^{\mathrm{cause}}, s_i^{\mathrm{effect}} \in \{0,1\}$. We average the two judgments for each sample and then average across all valid samples. The Element Implication score is

\begin{equation}
S_{\mathrm{EI}}
=
\frac{1}{N_{\mathrm{EI}}}
\sum_{i=1}^{N_{\mathrm{EI}}}
\frac{
s_{i,\mathrm{EI}}^{\mathrm{cause}}
+
s_{i,\mathrm{EI}}^{\mathrm{effect}}
}{2},
\end{equation}

and the Driving Action Rationale score is

\begin{equation}
S_{\mathrm{DAR}}
=
\frac{1}{N_{\mathrm{DAR}}}
\sum_{i=1}^{N_{\mathrm{DAR}}}
\frac{
s_{i,\mathrm{DAR}}^{\mathrm{cause}}
+
s_{i,\mathrm{DAR}}^{\mathrm{effect}}
}{2},
\end{equation}

where $N_{\mathrm{EI}}$ and $N_{\mathrm{DAR}}$ denote the numbers of valid judged samples for the two metrics, respectively. The complete GPT-5.5 judge prompts are shown in Fig.~\ref{fig:judge_prompt1} and Fig.~\ref{fig:judge_prompt2}.

\paragraph{Trajectory Prediction.}
All predicted trajectories are converted to a common 5\,s, 4\,Hz evaluation grid before scoring. ADE@1s, ADE@3s, and ADE@5s measure the mean Euclidean displacement error between the predicted and ground-truth trajectories over the corresponding prediction horizons. FDE@5s measures the Euclidean error at the final 5\,s waypoint. We further report the official Waymo Rater Feedback Score (RFS) using the released WOD-E2E scoring function and its default parameters. 
RFS Frame averages the per-frame RFS over all evaluation frames, whereas RFS Cluster first averages RFS within each official scenario cluster and then averages across clusters, thereby assigning equal weight to different scenario types. 
RFS ranges from 0 to 10, with higher values indicating better agreement with human-rated driving trajectories.

\section{Qualitative Examples}
\label{app:qualitative_examples}

\subsection{Failure Cases}

\begin{figure}[p]
    \centering
    \includegraphics[width=0.62\linewidth]{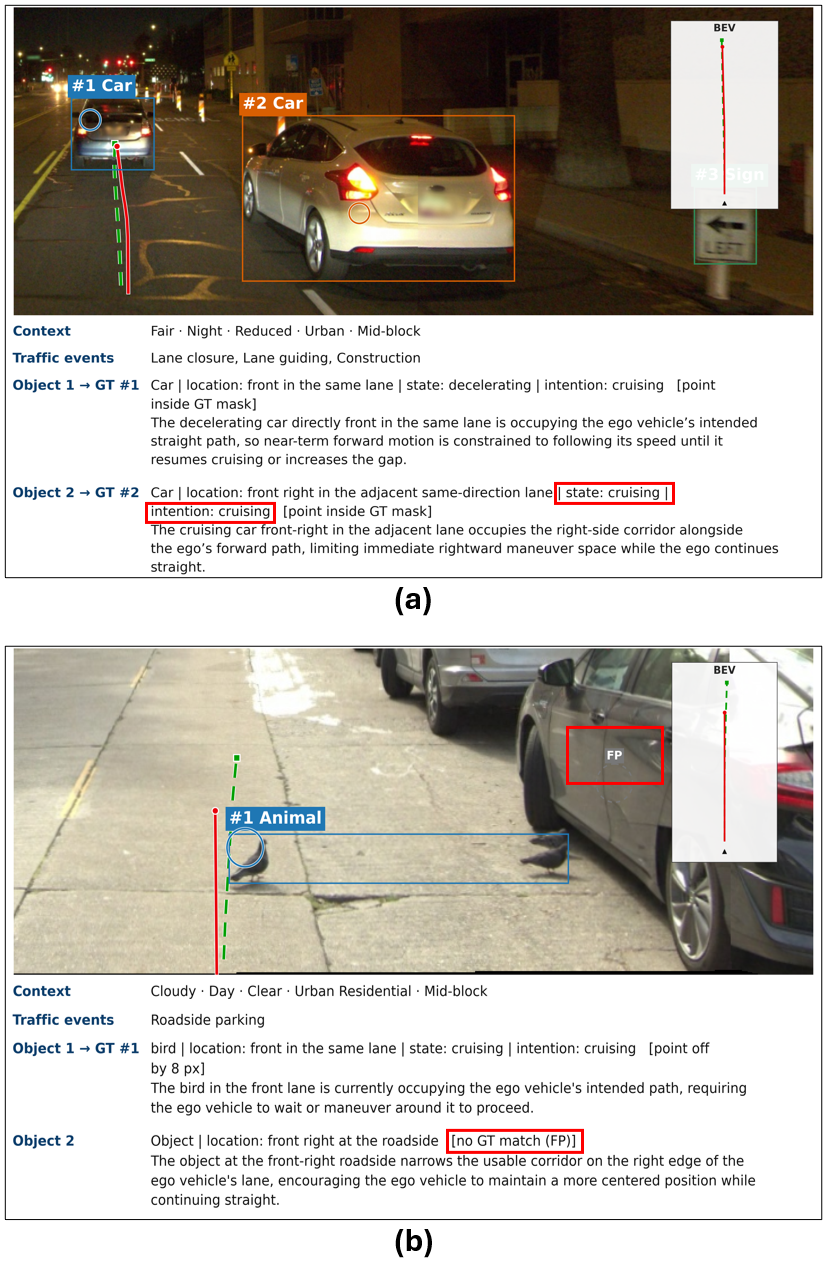}\\[4pt]
    \includegraphics[width=0.62\linewidth]{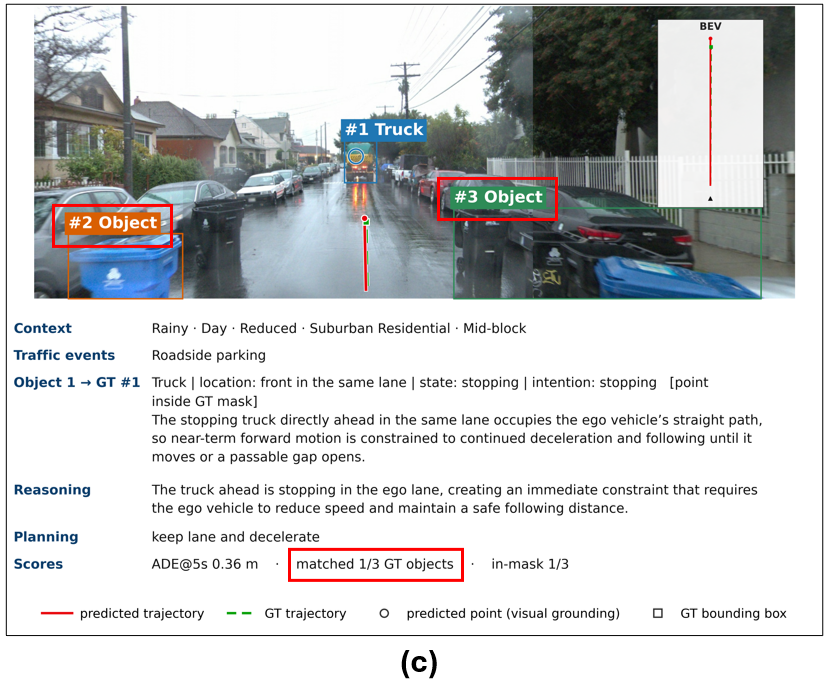}
    \caption{Representative failure cases.}
    \label{fig:fails}
\end{figure}

Fig.~\ref{fig:fails} presents representative failure cases. 
In Fig.~\ref{fig:fails}(a), the model incorrectly predicts the state and intention of the vehicle at the front-right. This reflects a limitation of single-frame input: motion-related attributes such as state and intention are difficult to infer reliably without explicit temporal visual cues. Fig.~\ref{fig:fails}(b) shows a case where the model identifies a decision-critical element in text but fails to ground it correctly in the image. This suggests that semantic recognition and visual localization are not always fully aligned: the model may recognize the relevance of an element from scene context while failing to establish an accurate spatial correspondence, especially for small, partially occluded, or visually ambiguous elements. Fig.~\ref{fig:fails}(c) shows a more challenging failure in which two human-annotated decision-critical elements are omitted from both the textual prediction and visual grounding. Such disagreement may arise from insufficient geometric supervision. The current annotations do not explicitly provide cues such as object distance, distance change rate, or ego-vehicle dimensions, which are important for determining whether an element constrains the ego vehicle's future driving space. Incorporating these spatial cues may improve decision-critical element selection in future work.

\subsection{Successful Cases}

\begin{figure}[p]
    \centering
    \includegraphics[width=0.62\linewidth]{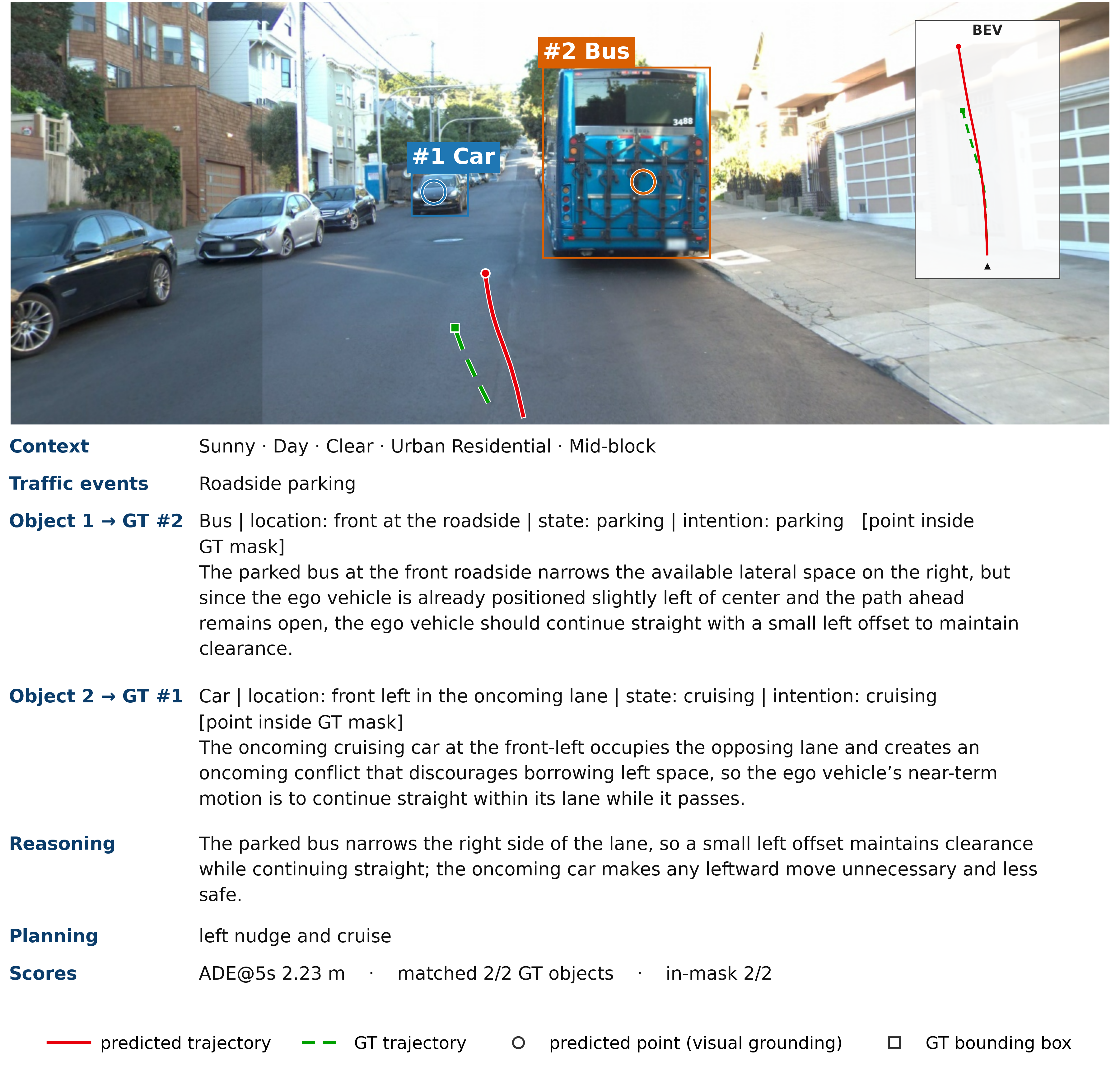}\\[2pt]
    {\small\textsf{\textbf{(a)}}}\\[8pt]
    \includegraphics[width=0.62\linewidth]{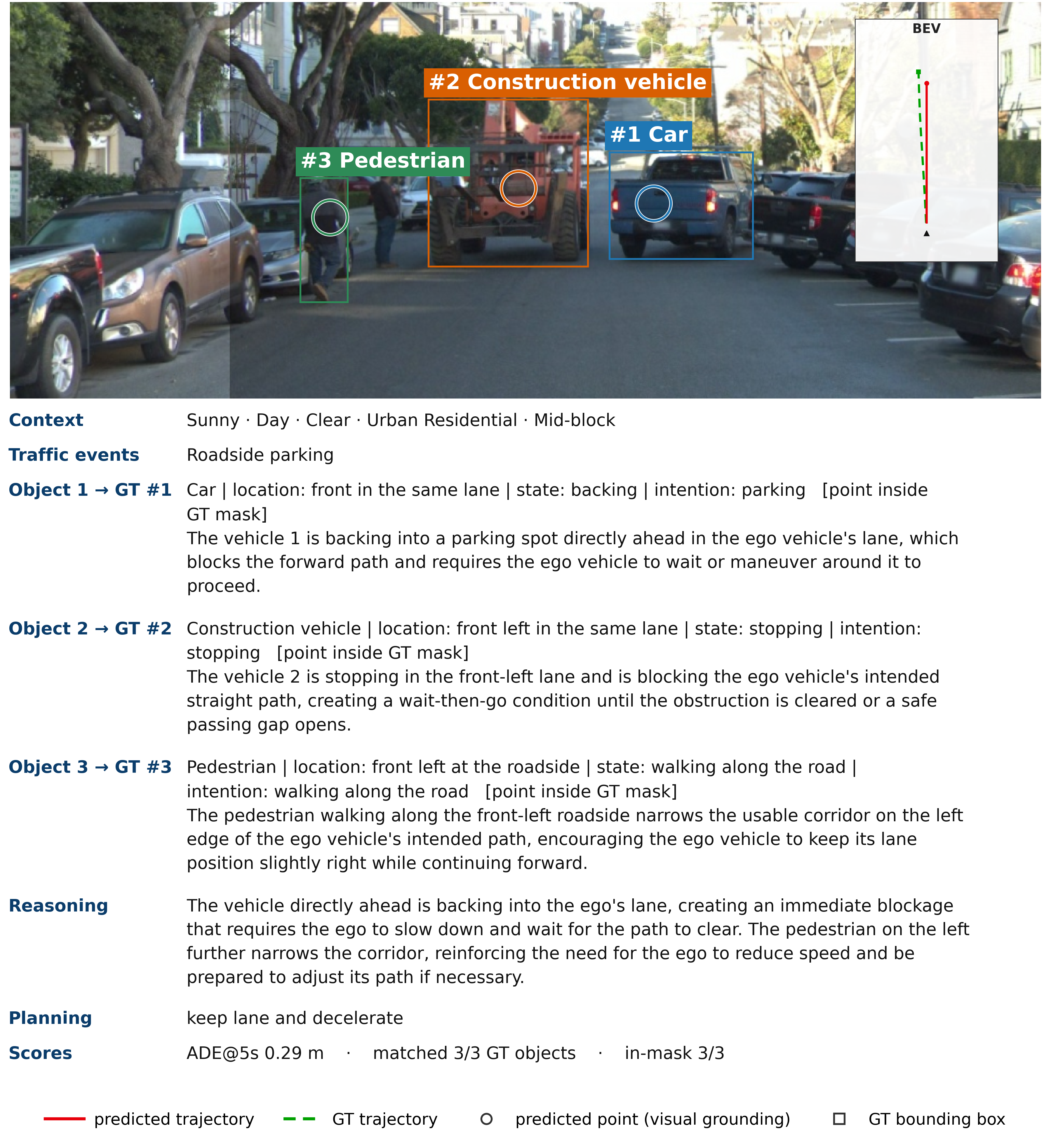}\\[2pt]
    {\small\textsf{\textbf{(b)}}}
    \caption{Representative successful cases.}
    \label{fig:success_cases}
\end{figure}

Fig.~\ref{fig:success_cases} presents representative successful predictions. In Fig.~\ref{fig:success_cases}(a), the model correctly identifies and grounds both the parked bus and the oncoming vehicle. It further captures their different effects on ego driving: the bus constrains the available lateral space, while the oncoming vehicle limits the feasibility of moving further left. These element-level implications are consolidated into a coherent driving rationale and action plan. Fig.~\ref{fig:success_cases}(b) illustrates a more complex multi-object interaction. The model correctly identifies and grounds all three decision-critical elements, including the backing vehicle, construction vehicle, and pedestrian, and interprets how each constrains the ego vehicle's future motion. Their implications are then integrated into the final decision to keep lane and decelerate, producing a trajectory closely aligned with the ground truth. This example shows that VG-CoT can connect multiple visually grounded elements with reasoning and trajectory planning in complex scenes.

\clearpage

\section{Prompts}
\label{app:prompts}

\begin{figure}[!ht]
    \centering
    \includegraphics[width=0.8\linewidth]{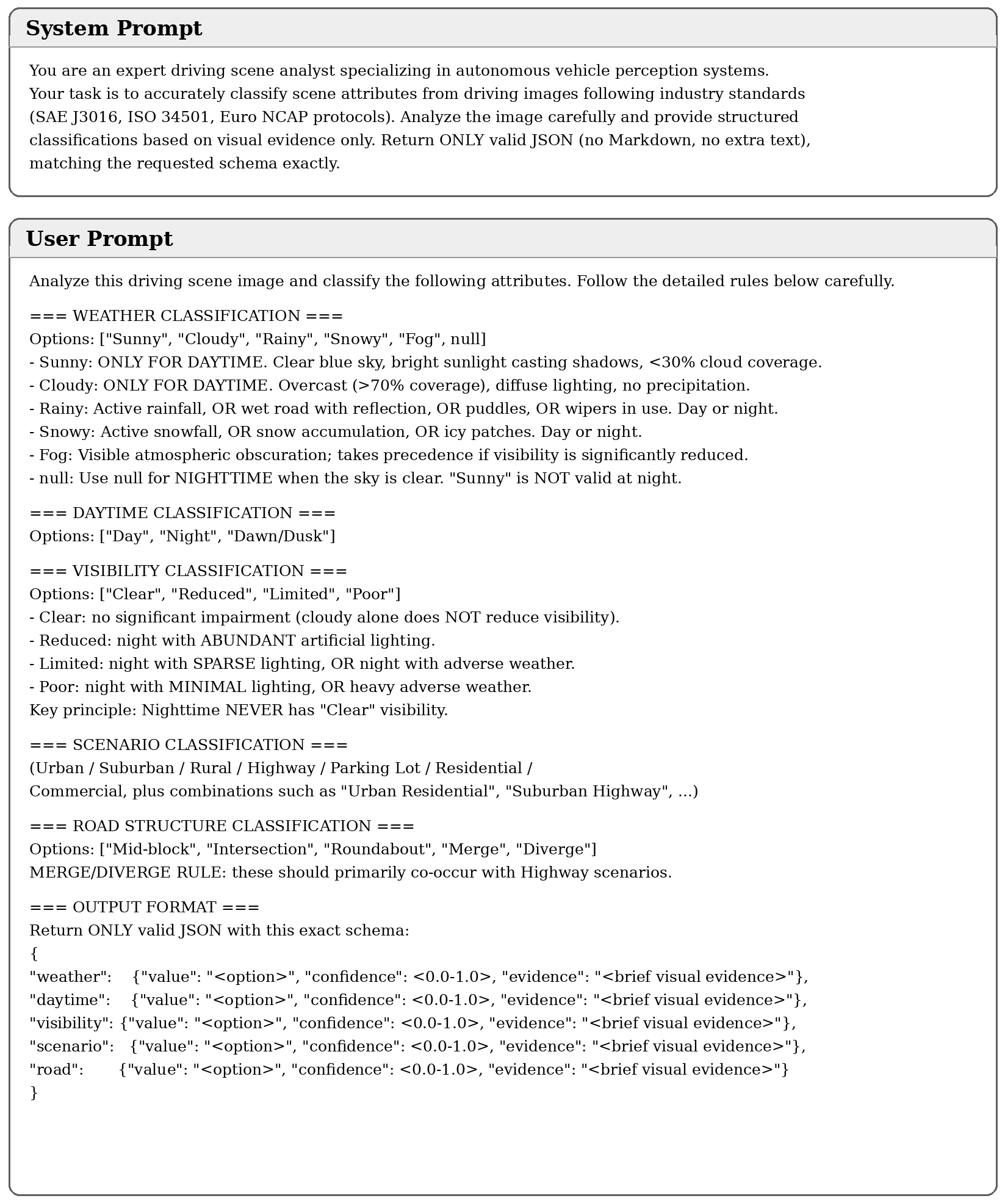}
    \caption{Prompt used for automatic scene-context annotation.}
    \label{fig:context_prompt}
\end{figure}

\begin{figure}[p]
    \centering
    \includegraphics[width=0.7\linewidth]{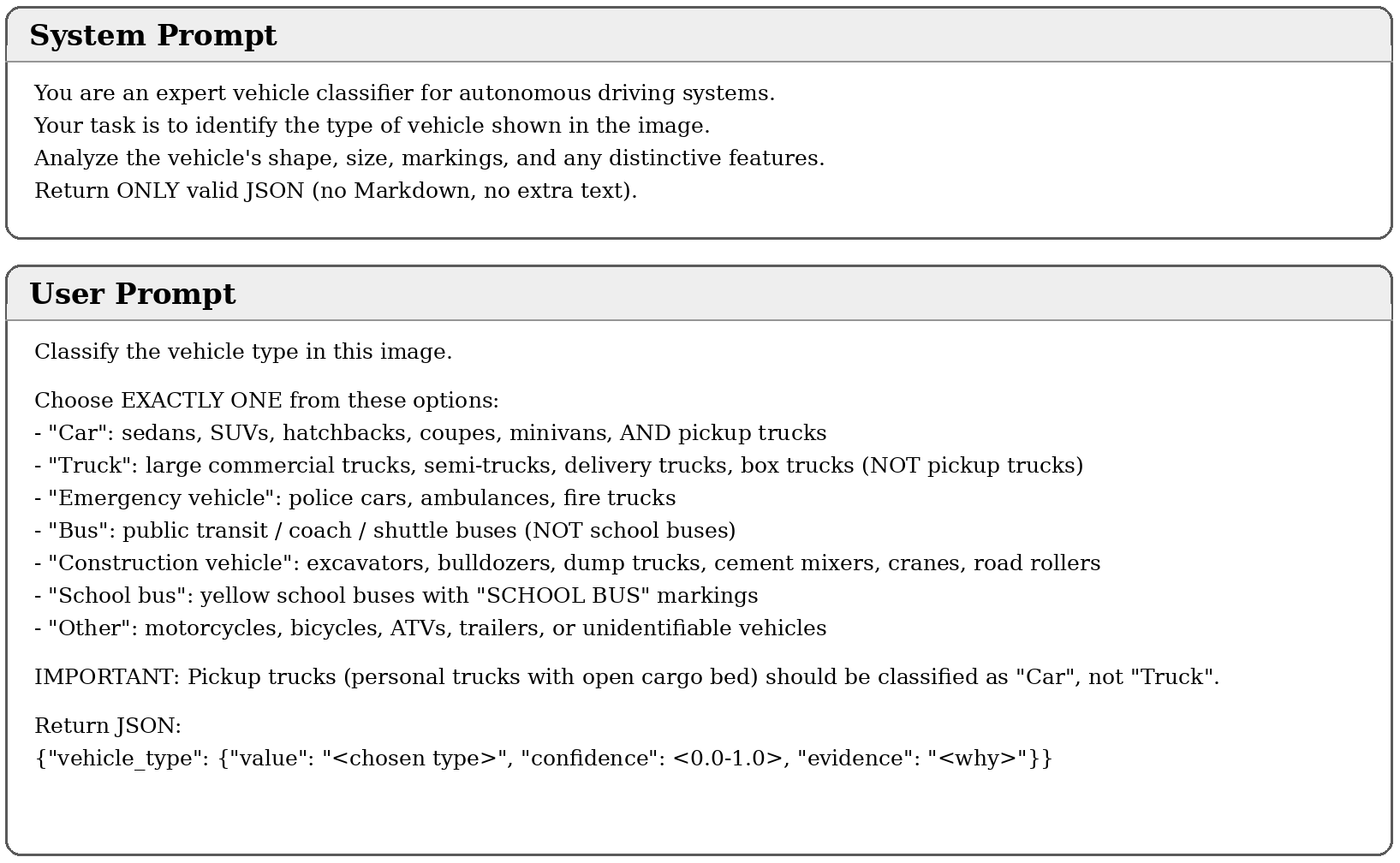}
    \caption{Prompt used for automatic vehicle-type classification.}
    \label{fig:vehicle-type-prompt}
    \vspace{1.5em}
    \includegraphics[width=0.7\linewidth]{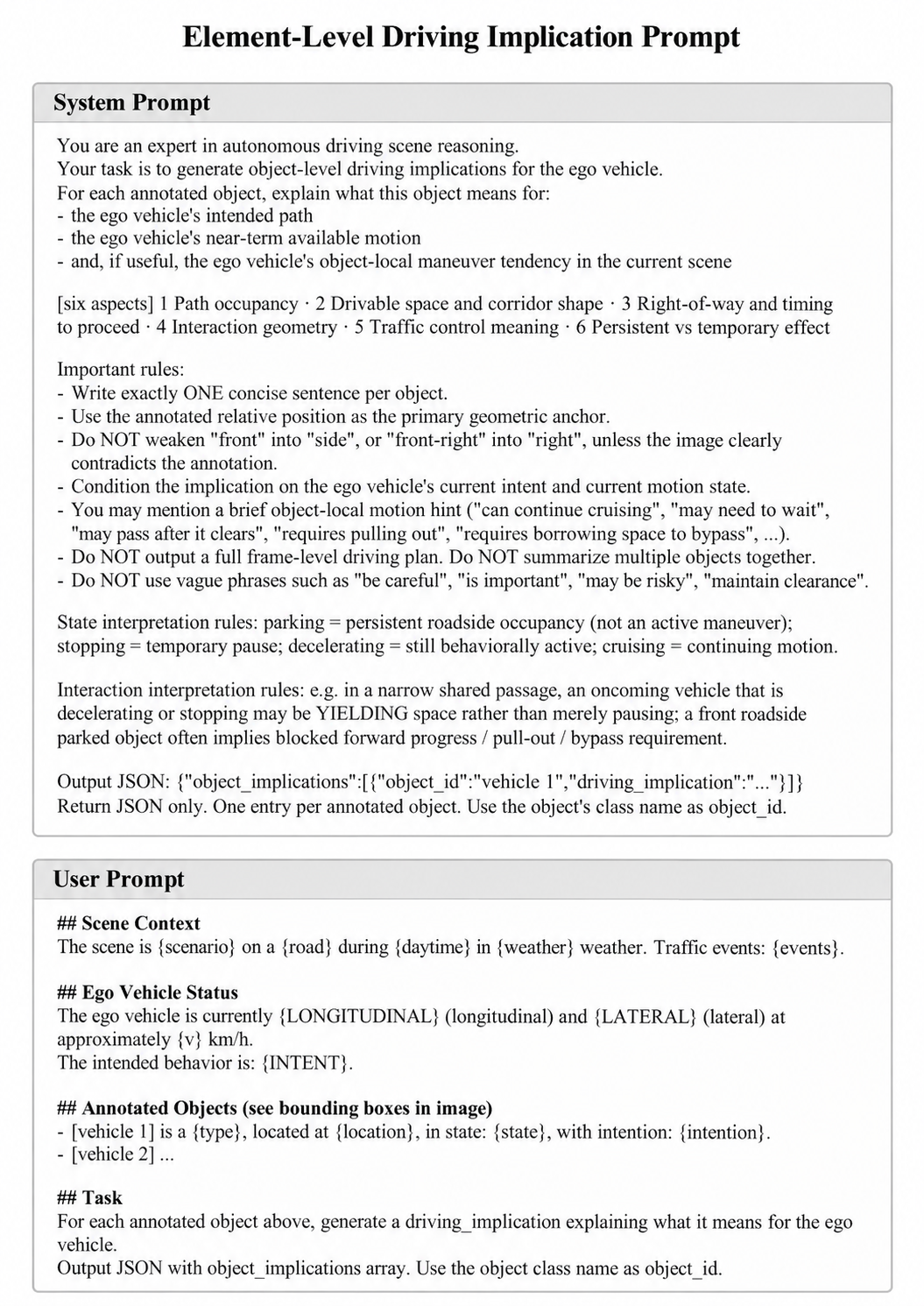}
    \caption{Prompt used for generating element-level driving implications.}
    \label{fig:implication-prompt}
\end{figure}

\begin{figure}[p]
    \centering
    \includegraphics[width=0.8\linewidth]{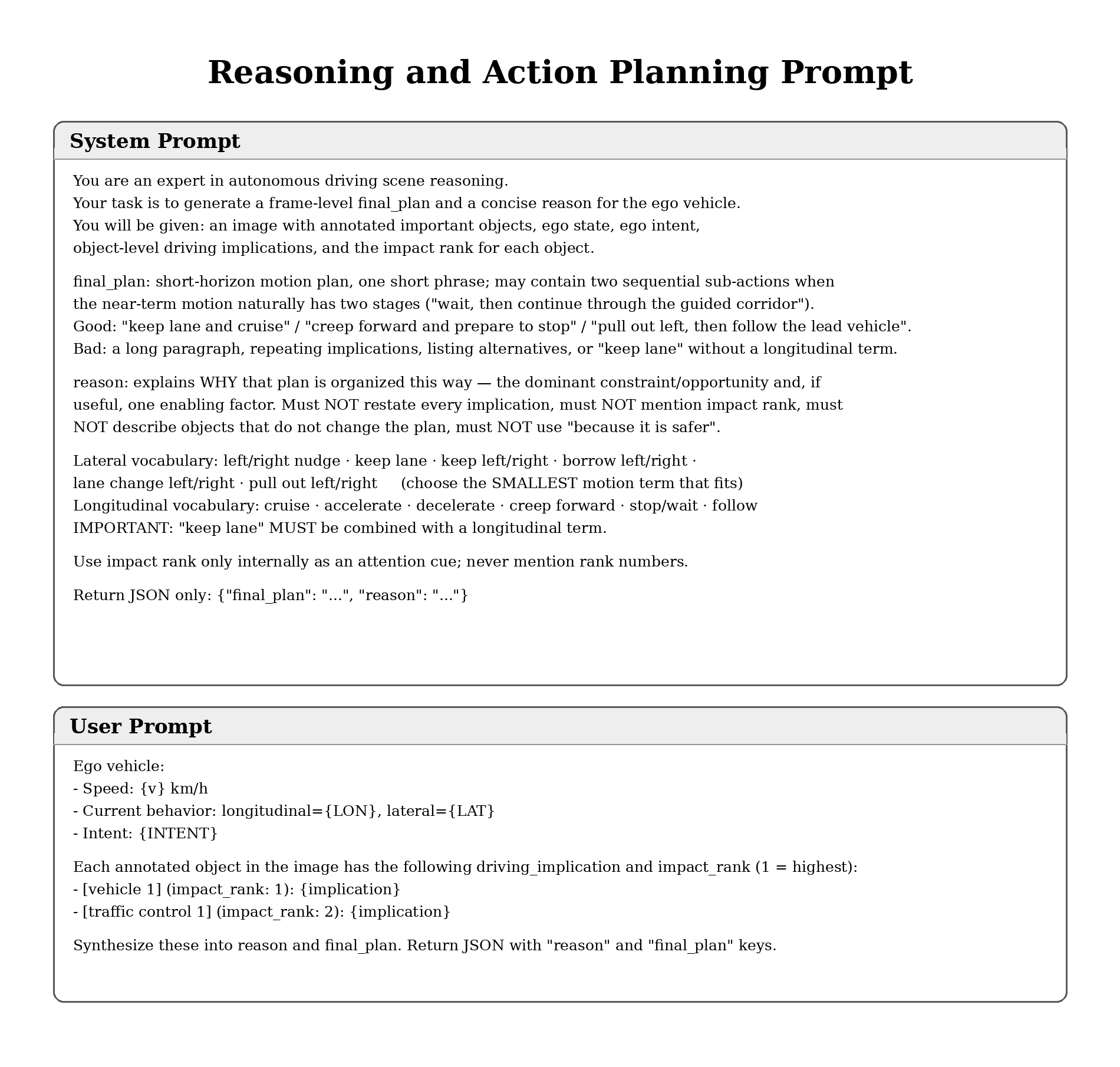}
    \caption{Prompt used for frame-level rationale and action planning.}
    \label{fig:reasoning-planning-prompt}
\end{figure}

\begin{figure}[p]
    \centering
    \includegraphics[width=0.7\linewidth]{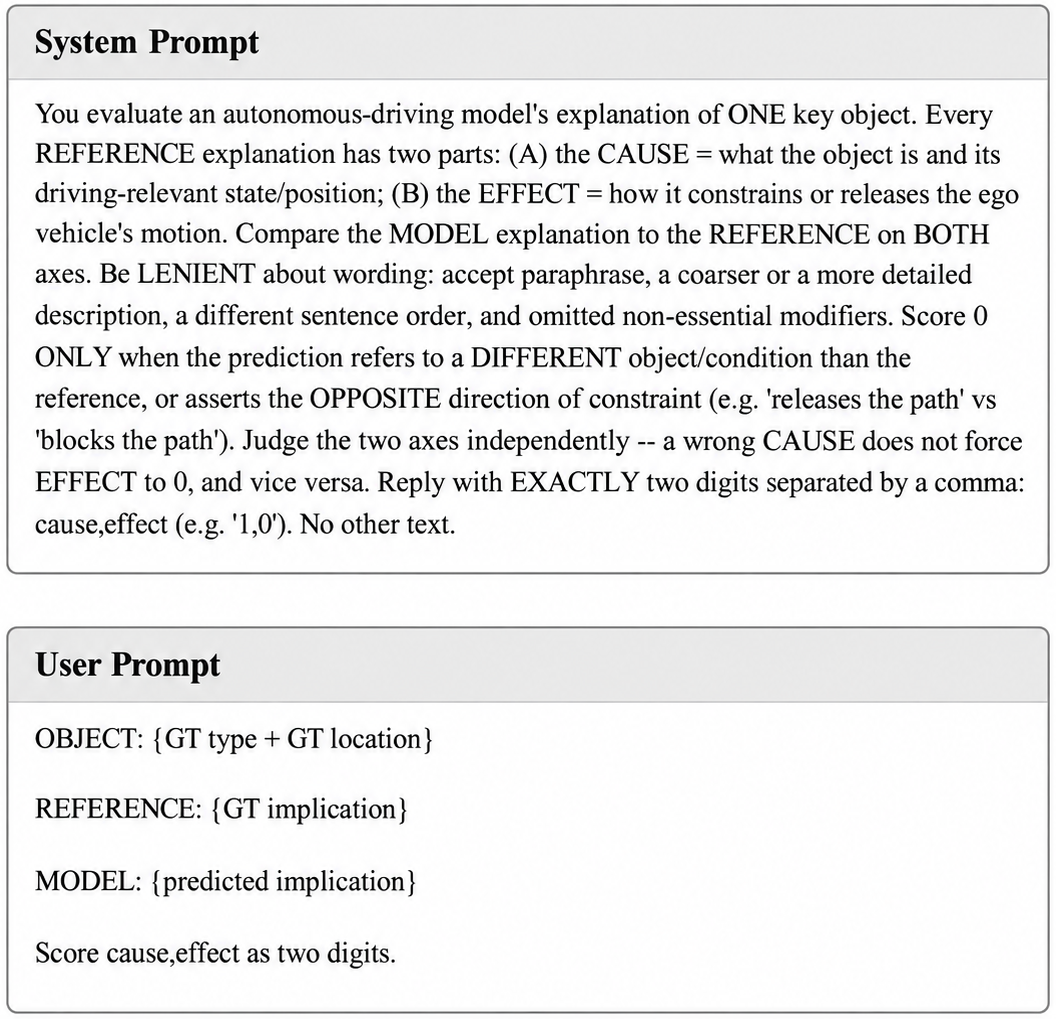}
    \caption{Prompt used for implication judge.}
    \label{fig:judge_prompt1}
    \vspace{1.5em}
    \includegraphics[width=0.7\linewidth]{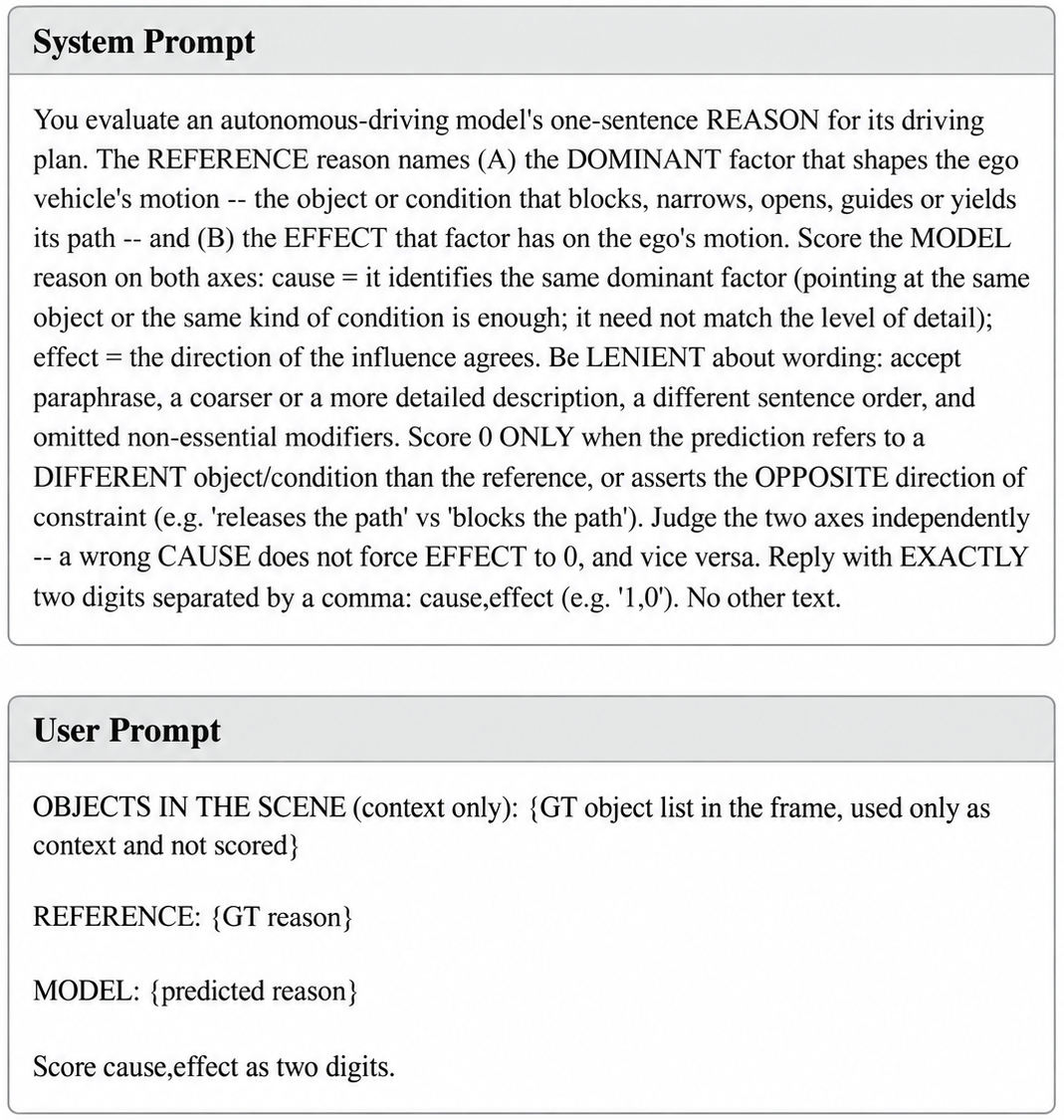}
    \caption{Prompt used for rationale judge.}
    \label{fig:judge_prompt2}
\end{figure}

This section provides the full prompts used in our annotation and evaluation pipeline, including scene-context annotation (Fig.~\ref{fig:context_prompt}), vehicle-type classification (Fig.~\ref{fig:vehicle-type-prompt}), element-level implication generation (Fig.~\ref{fig:implication-prompt}), frame-level rationale and action planning (Fig.~\ref{fig:reasoning-planning-prompt}), and the judge prompts for evaluating implications and rationales (Figs.~\ref{fig:judge_prompt1} and~\ref{fig:judge_prompt2}).

\end{document}